%% file: main.tex
\documentclass[10pt,twocolumn,letterpaper]{article}

\usepackage[pagenumbers]{cvpr} 

\input{preamble}

\definecolor{cvprblue}{rgb}{0.21,0.49,0.74}
\usepackage[pagebackref,breaklinks,colorlinks,allcolors=cvprblue]{hyperref}
\hypersetup{
  pdftitle={SPARE: Stage-Anchored Primitive Absorption and Residual Expansion for Lifelong VLAs},
  pdfauthor={Zhenlin He, Haoming Xu, Mingyue Li, Lei Lei, Yaqing Xu, Jiafeng Xu, Hao Dong},
}

\def\confName{CVPR}
\def\confYear{2027}

\title{SPARE: Stage-Anchored Primitive Absorption and Residual Expansion for Lifelong VLAs}

\author{
Zhenlin He\textsuperscript{1,2}\thanks{Equal contribution.} \quad
Haoming Xu\textsuperscript{1,3}\footnotemark[1] \quad
Mingyue Li\textsuperscript{2} \quad
Lei Lei\textsuperscript{4,5} \\
Yaqing Xu\textsuperscript{1,6} \quad
Jiafeng Xu\textsuperscript{2} \quad
Hao Dong\textsuperscript{1,2}\thanks{Corresponding author.}\\[0.5em]
\textsuperscript{1}QYRO \quad
\textsuperscript{2}Peking University\\
\textsuperscript{3}University of Chinese Academy of Sciences \quad
\textsuperscript{4}Shanghai Innovation Institute\\
\textsuperscript{5}University of Science and Technology of China \quad
\textsuperscript{6}Renmin University of China
}

\begin{document}
\maketitle
\input{sec/0_abstract}
\input{sec/1_intro}
\input{sec/2_related}
\input{sec/3_method}
\input{sec/4_experiments}
\input{sec/5_conclusion}
\input{sec/6_llm}
{
    \small
    \bibliographystyle{ieeenat_fullname}
    \bibliography{main}
}

\appendix
\input{sec/X_suppl}

\end{document}

%% file: preamble.tex
\usepackage[T1]{fontenc}

\usepackage{microtype}

\newcommand{\method}{SPARE\xspace}

%% file: sec/0_abstract.tex
\begin{abstract}
New manipulation tasks rarely require entirely new skills; instead, they recombine familiar action primitives while introducing only a few unseen ones. 
However, existing continual manipulation policies update uniformly at the task or stage level, ignoring this compositional structure and forcing a trade-off between catastrophic forgetting and restricted plasticity. 
We present \textbf{\method}, a primitive-aware residual mixture-of-experts policy that \emph{absorbs} familiar primitives into existing action experts and \emph{expands} capacity guided by kinematic novelty. 
\method assigns demonstration segments to experts in a self-supervised manner via kinematic novelty detection and error-driven competition. 
Because updating the shared backbone shifts the representations fed into inherited specialists, \method anchors expert-facing features and output velocities to the previous-stage policy to keep reused experts effective while mitigating forgetting.
On the official RoboCasa lifelong benchmark, \method outperforms continual-learning and replay baselines in cumulative success and final-stage skill retention.
On a continual benchmark we construct from RoboTwin~2.0, \method improves stage-2 retention after stage~3 by $15.4\%$ and zero-shot composition by $6.0\%$ over experience replay with distillation. 
On the dual-arm robot G2A, we collect demonstrations and train across five tasks over three stages. 
\method reaches $44.8\%$ overall success, outperforming replay with distillation by $16.8\%$, with gains of $18.0\%$ in stage-2 retention and $12.0\%$ in final-stage acquisition. 
These results demonstrate that adapting at the primitive level scales continual robot manipulation across open-ended task streams. Our code, benchmark splits, and post-trained checkpoints will be open-sourced.
\end{abstract}

%% file: sec/1_intro.tex
\section{Introduction}
\label{sec:intro}

\begin{figure*}[!t]
\centering
\includegraphics[width=0.93\textwidth]{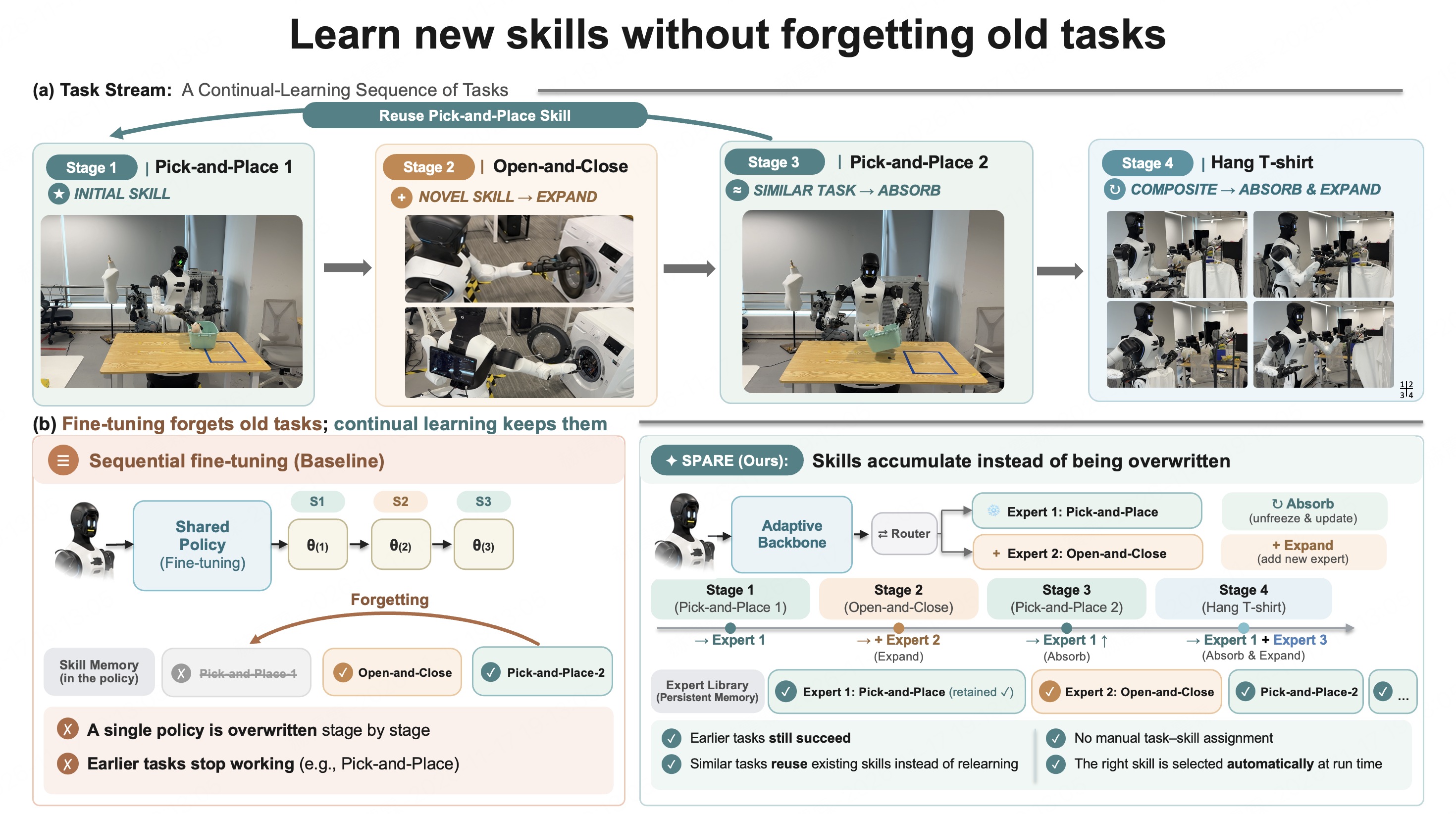}
\caption{\textbf{Motivation.} (a) A continual task stream mixes familiar primitives with unseen interactions. (b) Sequential fine-tuning overwrites earlier skills, whereas \method accumulates them: familiar segments are \emph{absorbed} into existing experts, capacity \emph{expands} guided by kinematic novelty, and experts are routed directly from observations at deployment.}
\label{fig:motivation}
\end{figure*}

Generalist manipulation policies, especially Vision-Language-Action (VLA) models~\cite{chi2023diffusionpolicy,octo2024,brohan2023rt2,kim2024openvla,black2025pi0,intelligence2025pi05}, learn broad action priors, yet real-world deployment requires continual adaptation to new tasks.
Naively fine-tuning these policies on sequential tasks overwrites earlier behaviors and causes severe catastrophic forgetting~\cite{wang2024clsurvey,yadav2026retain,zhu2026realworld}.
Instead, lifelong manipulation demands policies that acquire new skills while keeping prior ones intact and reusable~\cite{liu2023libero,zhu2026realworld} (\cref{fig:motivation}).

Existing continual manipulation methods follow two main directions.
Continual reinforcement learning (CRL) adapts policies through online interaction~\cite{zeng2026crlvla,hu2026simplerecipe}, which requires task-specific rewards and costly physical rollouts~\cite{lu2025vlarl,li2025simplevlarl,liu2025rl4vla}.
Continual imitation learning (CIL) learns from sequential demonstrations via experience replay and distillation~\cite{chen2026phaser,zhu2026realworld,zhao2026infovla,he2026lifelongvla}, skill libraries~\cite{zhang2026atomicvla,wu2025stellar}, or parameter-efficient adapters~\cite{liu2024tail,lei2025dmpel,romer2026clare}.
Both lines of work face a core stability--plasticity dilemma in the shared backbone.
Freezing the backbone preserves old skills but fails to ground unfamiliar interactions; updating it causes representation drift that compounds across stages~\cite{zeng2026crlvla,he2026lifelongvla,zhao2026infovla,yu2026hidra,romer2026clare}.
Resolving this trade-off requires rethinking \emph{where} and \emph{when} to allocate model capacity as the skill repertoire grows.

Our key insight comes from the compositional nature of manipulation: a new task is rarely entirely new.
Most tasks recombine familiar action primitives---such as grasping, placing, and pressing---while introducing only a few unseen ones.
For example, hanging a mug reuses grasping and transport before executing a new hanging motion.
Existing continual methods ignore this granularity and update uniformly across an entire stage.
Expansion-based methods attach stage-isolated modules to frozen backbones~\cite{liu2024tail,lei2025dmpel,romer2026clare}, blocking subtask reuse across stages, while fixed-capacity methods force diverse motions through a single shared policy~\cite{chen2026phaser,zhu2026realworld,wu2025stellar}.
In contrast, we align adaptation with primitive novelty: \emph{absorbing} familiar primitives into existing experts while \emph{expanding} capacity guided by kinematic novelty.

To instantiate this idea, we present \textbf{\method}, a primitive-aware residual mixture-of-experts (MoE) policy for continual manipulation.
\method pairs a pretrained generalist policy with a growing library of residual action experts that predict primitive-specific refinements.
For incoming demonstrations, \method uses kinematic clustering against prior primitive families to automatically decide how many experts to add: familiar motions refine existing specialists, whereas novel kinematic clusters spawn new residual experts.
During training, subtask segments are assigned to experts in a self-supervised manner via error-driven competition.
At inference, a lightweight chunk-level router dispatches experts directly from visual-language observations, enabling compositional reuse across multi-step tasks.
Crucially, expert reuse requires a stable representation interface: even if prior experts are frozen, updating the shared backbone shifts the features fed into the router and inherited experts, rendering prior specialists ineffective.
\method therefore keeps the shared backbone adaptable while anchoring expert-facing features and output velocities to the previous-stage policy, making continual expert reuse feasible while mitigating forgetting.

Our main contributions are threefold:
\begin{itemize}\setlength{\itemsep}{1pt}\setlength{\parskip}{0pt}\setlength{\topsep}{2pt}
  \item We propose \textbf{\method}, a primitive-aware residual MoE policy that selectively absorbs familiar primitives or expands capacity guided by kinematic novelty detection and self-supervised error competition.
  \item We introduce stage-wise feature and velocity anchoring as a prerequisite for expert reuse under an adaptable shared backbone, enabling new-task acquisition while mitigating interference on previously learned skills.
  \item We construct a $50$-task continual RoboTwin~2.0 benchmark and demonstrate across the official RoboCasa lifelong benchmark, RoboTwin~2.0, and five tasks on the dual-arm robot G2A that \method outperforms continual-learning and matched-replay baselines.
\end{itemize}

%% file: sec/2_related.tex
\section{Related work}
\label{sec:related}

\noindent\textbf{Generalist manipulation policies and VLAs.}
Diffusion and flow-matching policies~\cite{chi2023diffusionpolicy,lipman2023flow,octo2024,liu2025rdt,nvidia2025gr00t} as well as large-scale VLAs~\cite{brohan2023rt2,kim2024openvla,black2025pi0,intelligence2025pi05} learn broad action priors, yet sequential fine-tuning on downstream tasks~\cite{kim2025oft,lu2025vlarl,li2025simplevlarl,liu2025rl4vla} quickly overwrites earlier skills~\cite{yadav2026retain,zhu2026realworld}.
To preserve prior capabilities, RETAIN~\cite{yadav2026retain} linearly interpolates between pretrained and fine-tuned weights at each stage, but repeated merging progressively dilutes plasticity for new tasks.
VLA-MOPD~\cite{lv2026vlamopd} instead distills separately trained task specialists into one student policy, yet requires concurrent access to all task teachers rather than adapting sequentially.
Unlike weight merging or offline multi-teacher distillation, \method adapts sequentially by keeping a shared flow-matching generalist~\cite{intelligence2025pi05} and isolating primitive refinements in residual experts.

\smallskip\noindent\textbf{Continual robot imitation learning.}
Building on classical continual learning~\cite{kirkpatrick2017ewc,li2016lwf,rebuffi2017icarl,rusu2016pnn,wang2024clsurvey} and robot benchmarks~\cite{wolczyk2021continualworld,liu2023libero,zeng2026crlvla,hu2026simplerecipe,nasiriany2026robocasa365}, recent continual VLA methods fall into fixed-capacity and stage-isolated designs.
\mbox{LifelongVLA}~\cite{he2026lifelongvla}, InfoVLA~\cite{zhao2026infovla}, and PHASER~\cite{chen2026phaser} regularize internal features or combine experience replay with distillation inside a single policy; however, forcing diverse motions through one shared action head causes gradient conflict across stages.
Conversely, TAIL~\cite{liu2024tail} and DMPEL~\cite{lei2025dmpel} attach task- or stage-specific LoRA modules to a frozen backbone, while CLARE~\cite{romer2026clare} inserts expandable adapters routed by autoencoder reconstruction error.
Unlike these methods, which freeze the backbone and isolate modules by stage---preventing subtask reuse and limiting adaptation to unfamiliar interactions---\method allocates capacity at the primitive level and anchors an adaptable backbone.
Agentic or routing-based systems~\cite{tao2026prism,zhang2026harnessvla} instead compose frozen policies with memory-guided agents or in-context skills, whereas \method directly adapts low-level continuous control.

\smallskip\noindent\textbf{Mixture of experts and compositional skill reuse.}
Modular skill libraries~\cite{wan2024lotus,wu2025stellar,quang2026retrievetoact} and sparse MoEs~\cite{shazeer2017moe,fedus2022switch,aljundi2017expertgate,yu2024moeadapters,wang2024sdp} decompose complex behaviors into specialists.
In manipulation, AtomicVLA~\cite{zhang2026atomicvla} trains discrete skill modules scheduled by a high-level planner, whereas \method routes continuous action refinements end-to-end without hierarchical supervision.
Inside transformer blocks, AdaMoE~\cite{shen2025adamoe} scales the action expert with token-level MoE, and HIDRA~\cite{yu2026hidra} adds hierarchical dual-attention routing with an expandable expert set.
However, AdaMoE keeps a fixed expert pool, and token-wise gating in both breaks temporal coherence within action chunks.
\method instead pairs a generalist branch with dynamically expandable residual experts routed at the chunk level, enabling coherent execution and zero-shot composition.

%% file: sec/3_method.tex
\section{Method}
\label{sec:method}

\begin{figure*}[!t]
\centering
\includegraphics[width=0.91\textwidth]{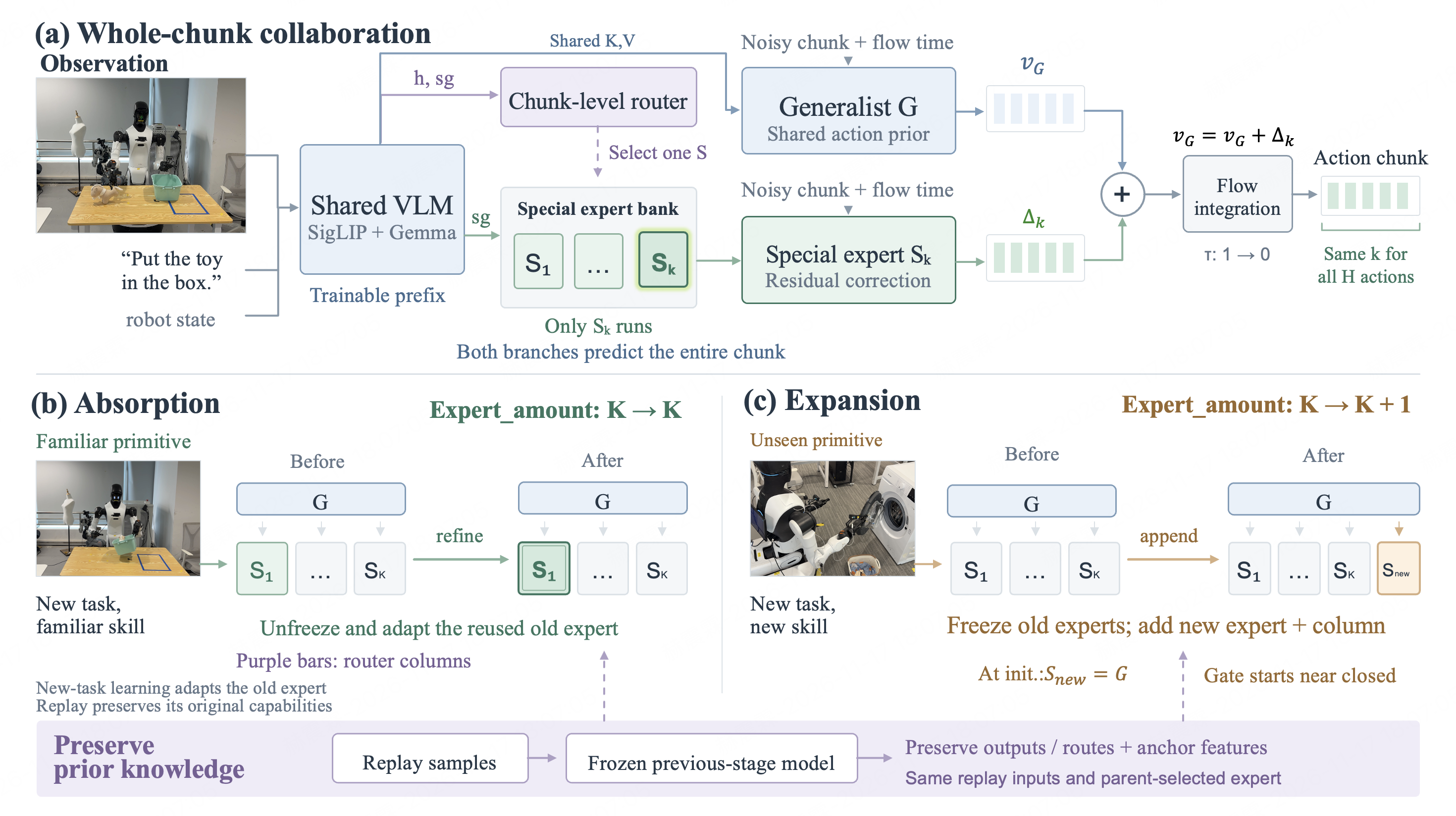}
\caption{\textbf{Overview of \method.} (a) Shared backbone $f_\phi$ produces prefix tokens $\mathbf{h}$ and caches $(\mathbf{K},\mathbf{V})$ to condition generalist $G$ ($g_\theta$, velocity $\mathbf{v}_G$) and residual experts $\{S_k\}_{k=1}^K$ ($\boldsymbol\Delta_k$), routed per chunk by $r_\omega$. (b) \emph{Absorption} refines matched experts at fixed $K$. (c) \emph{Expansion} freezes unmatched prior experts, appends $\{S_j\}_{j\in\mathcal{I}_{\mathrm{new}}}$, and anchors replay routes, velocities, and features to the frozen parent.}
\label{fig:method}
\end{figure*}

Built upon a shared generalist manipulation policy, \method addresses continual adaptation at the primitive level (\cref{fig:method}). It organizes reusable motion primitives into residual action experts, \emph{absorbs} familiar primitives by refining existing specialists, and \emph{expands} capacity guided by kinematic novelty, while anchoring shared backbone features and output velocities across stages.

\subsection{Continual learning with a flow-based policy}
\label{sec:setup}
We consider $S$ sequential training stages. Each stage $s\in\{1,\ldots,S\}$ provides a demonstration dataset $\mathcal{D}_s$ over tasks $\mathcal{T}_s$, where each training chunk $c=(o,\ell,\mathbf{A})\in\mathcal{D}_s$ consists of multi-view RGB and proprioceptive observation $o$, language instruction $\ell$, and target action chunk $\mathbf{A}\in\mathbb{R}^{H\times d}$ with horizon $H$ and action dimension $d$. After stage $s$, a single policy is evaluated across all seen tasks $\mathcal{T}_{\le s}=\bigcup_{i=1}^{s}\mathcal{T}_i$ directly from $(o,\ell)$. For $s\ge2$, a fraction $\rho\in(0,1)$ of each mini-batch is sampled from a replay buffer $\mathcal{R}$ containing demonstrations from earlier stages.

We build upon $\pi_{0.5}$~\cite{intelligence2025pi05}. Its vision-language backbone $f_\phi$ encodes $(o,\ell)$ into prefix tokens $\mathbf{h}$ and layer-wise key-value caches $\mathbf{K}=\{\mathbf{K}_i\}_{i=1}^L$, $\mathbf{V}=\{\mathbf{V}_i\}_{i=1}^L$, conditioning the generalist action branch $G$ ($g_\theta$). Given flow time $\tau\sim\mathrm{Beta}(1.5,1)$ and Gaussian noise $\boldsymbol\epsilon\sim\mathcal{N}(\mathbf{0},\mathbf{I})$ in $\mathbb{R}^{H\times d}$, we form $\mathbf{A}^{\tau}=\tau\boldsymbol\epsilon+(1-\tau)\mathbf{A}$ and target velocity $\mathbf{u}=\boldsymbol\epsilon-\mathbf{A}$, training via flow matching~\cite{lipman2023flow}:
\begin{equation}
\mathcal{L}_{\mathrm{FM}}=\mathbb{E}_{c\sim\mathcal{D}_s,\tau,\boldsymbol\epsilon}\Big\|g_\theta(\mathbf{A}^{\tau},\tau\mid f_\phi(o,\ell))-\mathbf{u}\Big\|_F^2,
\label{eq:fm}
\end{equation}
with Frobenius norm $\|\cdot\|_F$. At inference, fixed-step Euler integration evolves predicted velocities from $\tau=1$ to $\tau=0$.

\subsection{Reusable residual action experts}
\label{sec:moe}
\noindent\textbf{Generalist--expert collaboration.}
Instead of replacing the pretrained action head, we keep the generalist velocity $\mathbf{v}_G=g_\theta(\mathbf{A}^{\tau},\tau\mid f_\phi(o,\ell))\in\mathbb{R}^{H\times d}$ as a shared prior and maintain a library $\mathcal{E}=\{S_1,\ldots,S_K\}$ of $K$ residual action experts ($E_0$--$E_{K-1}$ in routing plots). Each expert $S_k$ ($\delta_{\psi_k}$) attends to $(\mathbf{K},\mathbf{V})$ to predict a primitive-specific refinement $\boldsymbol\Delta_k=\delta_{\psi_k}(\mathbf{A}^{\tau},\tau\mid f_\phi(o,\ell))\in\mathbb{R}^{H\times d}$, producing composite velocity $\mathbf{v}=\mathbf{v}_G+\boldsymbol\Delta_k$. Initializing $\psi_k$ from $\theta$ with small weight perturbations and a zero-initialized output projection ensures every expert starts at $\mathbf{v}_G$ and learns only residual refinements.

\smallskip\noindent\textbf{Chunk-level routing.}
A lightweight router $r_\omega$ applies attention pooling over stop-gradient prefix tokens $\mathrm{sg}(\mathbf{h})$, which keeps early routing noise from perturbing $f_\phi$, to extract router feature $\mathbf{x}\in\mathbb{R}^{d_r}$, and linearly maps $\mathbf{x}$ to expert logits $\mathbf{z}=(z_1,\ldots,z_K)^\top\in\mathbb{R}^{K}$. Selecting $k^{\star}=\arg\max_{k\in\{1,\ldots,K\}} z_k$ once prior to Euler integration executes a single residual expert across all $H$ steps, preserving intra-chunk temporal coherence and keeping inference cost independent of $K$, while multi-step tasks switch experts across successive chunks from $(o,\ell)$.

\subsection{Bootstrapping a primitive expert library}
\label{sec:stage1}
In Stage~1, we initialize $K$ residual experts on $\mathcal{D}_1$ by minimizing $\|\mathbf{v}_G+\boldsymbol\Delta_{k(c)}-\mathbf{u}\|_F^2$ for each chunk $c\in\mathcal{D}_1$ and its assigned expert index $k(c)\in\{1,\ldots,K\}$. To avoid early routing collapse, $k(c)$ is briefly warmed up via $K$-means clustering over standardized trajectory kinematics (net end-effector translation, cumulative rotation, active-arm masks, and gripper open-close profiles). Thereafter, assignment becomes self-supervised and error-driven: we evaluate candidate flow losses across all $K$ experts for each chunk $c$, and reassign $k(c)$ only if another expert lowers the loss by at least $10\%$. Cross-entropy loss supervises $r_\omega$ to predict $k(c)$ while only $g_\theta$ backpropagates into $f_\phi$. After Stage~1, we record the kinematic cluster collection $\mathcal{P}_1=\{1,\ldots,K\}$, storing each cluster $m$'s mean $\boldsymbol\mu_m$, diagonal covariance $\boldsymbol\Sigma_m$, $95$th-percentile intra-cluster distance threshold $\gamma_m$, and matched expert index $a(m)\in\{1,\ldots,K\}$.

\subsection{Primitive-aware absorption and expansion}
\label{sec:absorb}
\noindent\textbf{Kinematic novelty detection.}
Here, each primitive cluster represents a reusable kinematic motion pattern (e.g., single-arm pick-and-place or bimanual lifting) rather than a task label. Given $\mathcal{D}_s$ in stage $s\ge2$, we partition demonstrations into candidate segments $\{c_i\}$ at gripper transitions or via sliding $H$-frame windows, and extract each segment's kinematic descriptor $\boldsymbol\phi(c_i)\in\mathbb{R}^{d_\phi}$ from the same four Stage~1 kinematic features. Let $\mathcal{P}_{<s}=\{1,\ldots,|\mathcal{P}_{<s}|\}$ index all primitive clusters accumulated before stage $s$, with stored centroids $\{\boldsymbol\mu_m\}$, diagonal covariances $\{\boldsymbol\Sigma_m\}$, thresholds $\{\gamma_m\}$, and expert mapping $a(m)\in\{1,\ldots,K\}$. For each segment $c_i$, we find its nearest prior cluster $m^{\star}$ and normalized Mahalanobis distance $d_{\min}(c_i)$:
\begin{equation}
d_{\min}(c_i)=\min_{m\in\mathcal{P}_{<s}}\big\|\boldsymbol\Sigma_m^{-1/2}\big(\boldsymbol\phi(c_i)-\boldsymbol\mu_m\big)\big\|_2,
\label{eq:novelty-dist}
\end{equation}
with minimizing cluster index $m^{\star}$. Comparing $d_{\min}(c_i)$ with threshold $\gamma_{m^{\star}}$ separates familiar motions from novel ones. Segments with $d_{\min}(c_i)\le\gamma_{m^{\star}}$ are classified as familiar and linked to existing expert $S_{a(m^{\star})}$. Outlier segments with $d_{\min}(c_i)>\gamma_{m^{\star}}$ are grouped via $k$-means, with the cluster count chosen by silhouette score: compact outlier clusters exceeding $5\%$ of stage-$s$ segments automatically register $M_s\ge 0$ novel primitive clusters, while smaller groups are treated as noise. This yields a selective adaptation rule for each segment $c_i$:
\begin{equation}
\mathcal{U}(c_i)=
\begin{cases}
\text{absorb into }S_{a(m^{\star})}, & d_{\min}(c_i)\le\gamma_{m^{\star}},\\[2pt]
\text{expand ($M_s$ experts)}, & \begin{aligned}[t]
  &d_{\min}(c_i)>\gamma_{m^{\star}},\\[-2pt]
  &M_s\ge 1.
\end{aligned}
\end{cases}
\label{eq:decision}
\end{equation}

When $M_s=0$, \method performs pure absorption without adding experts, initializing chunk targets from matched specialists $S_{a(m^{\star})}$ (or querying the frozen stage-$(s{-}1)$ router) and refining them via $10\%$-margin error competition across the $K$ existing experts. When $M_s\ge1$, \method instantiates $M_s$ new residual experts $\{S_j\}_{j\in\mathcal{I}_{\mathrm{new}}}$ indexed by $\mathcal{I}_{\mathrm{new}}=\{K{+}1,\ldots,K{+}M_s\}$, warms them up on the $M_s$ novel clusters, and runs $10\%$-margin error competition across all $K+M_s$ experts so that familiar segments within composite tasks route back to inherited specialists $S_{a(m^{\star})}$. After stage $s$, the $M_s$ new cluster statistics and expert mappings are appended to $\mathcal{P}_{<s}$, and $K\leftarrow K+M_s$ (with $\mathcal{I}_{\mathrm{new}}=\emptyset$ when $M_s=0$). At inference, routing relies solely on $(o,\ell)$ without kinematic segmentation.

\smallskip\noindent\textbf{Absorption: refining existing capacity.}
Let $\mathcal{I}_{\mathrm{abs}}\subseteq\{1,\ldots,K\}$ index prior experts matched to familiar segments in stage $s$. Unfreezing $\{S_k\}_{k\in\mathcal{I}_{\mathrm{abs}}}$ adapts them to new object geometries and spatial layouts, while replay velocity anchoring (\cref{sec:anchor}) mitigates drift on earlier tasks. In stages involving multi-primitive compositions, $g_\theta$ is jointly unfrozen to smooth transitions between primitives without adding parameters.

\smallskip\noindent\textbf{Expansion: allocating capacity via kinematic novelty.}
When $M_s\ge1$, we freeze unmatched prior experts $\{S_k\}_{k\in\{1,\ldots,K\}\setminus\mathcal{I}_{\mathrm{abs}}}$, i.e., all prior experts when $\mathcal{I}_{\mathrm{abs}}=\emptyset$, and append zero-output-initialized experts $\{S_j\}_{j\in\mathcal{I}_{\mathrm{new}}}$, while matched experts in $\mathcal{I}_{\mathrm{abs}}$ remain unfrozen in composite stages. Given frozen-column logits $z_1,\ldots,z_K$ and router feature $\mathbf{x}$, each expanded logit $z_j$ ($j\in\mathcal{I}_{\mathrm{new}}$) is parameterized as:
\begin{equation}
z_j=\log\!\sum_{k=1}^{K}\exp(z_k)+e_j(\mathbf{x}),\quad j\in\mathcal{I}_{\mathrm{new}},
\label{eq:ext}
\end{equation}
where $e_j(\mathbf{x})=\mathbf{w}_j^\top\mathbf{x}+b_j$ ($\mathbf{w}_j\in\mathbb{R}^{d_r}$, $b_j\in\mathbb{R}$). Adding a single expanded expert assigns it logistic probability $\sigma(e_j(\mathbf{x}))$ while preserving exact relative probabilities among all $K$ prior experts. Initializing $\mathbf{w}_j=\mathbf{0}$ and $b_j=-4$ caps each new branch's initial mass below $2\%$, mitigating early routing drift.

\smallskip\noindent\textbf{Stage-wise adaptation objective.}
On current-stage chunks $c\sim\mathcal{D}_s$, we minimize the action flow loss $\mathcal{L}_{\mathrm{act}}=\mathbb{E}_{c\sim\mathcal{D}_s,\tau,\boldsymbol\epsilon}\|\mathbf{v}_G+\boldsymbol\Delta_{k(c)}-\mathbf{u}\|_F^2$, where $k(c)\in\{1,\ldots,K{+}M_s\}$ is the assigned chunk expert. Over the full logit vector $\tilde{\mathbf{z}}=(z_1,\ldots,z_{K+M_s})^\top\in\mathbb{R}^{K+M_s}$, cross-entropy loss $\mathcal{L}_{\mathrm{route}}$ supervises $r_\omega$ toward $k(c)$ on $\mathcal{D}_s$ and toward the frozen parent router's top-1 expert $a_0\in\{1,\ldots,K\}$ on replay samples $c\sim\mathcal{R}$. Meanwhile, backbone $f_\phi$ remains trainable with a conservative learning rate across stages.

\subsection{Mitigating drift and forgetting via stage-wise anchoring}
\label{sec:anchor}
Isolating parameters in prior experts---even when frozen---is insufficient to curb forgetting once $f_\phi$ updates to ground new tasks, because shifts in $\mathbf{h}$ and $(\mathbf{K},\mathbf{V})$ alter the conditioning interface shared by $r_\omega$ and all inherited specialists; meanwhile, specialists unfrozen during absorption ($\mathcal{I}_{\mathrm{abs}}$) also require output regularization on prior tasks. We therefore retain a frozen copy of the stage-$(s{-}1)$ policy as an anchor on replay samples $c\sim\mathcal{R}$ evaluated at the same noisy action $(\mathbf{A}^{\tau},\tau)$.

\smallskip\noindent\textbf{Output velocity anchoring.}
For each replay sample $c\sim\mathcal{R}$ with parent-selected expert $a_0\in\{1,\ldots,K\}$, let $\mathbf{v}'=\mathbf{v}'_G+\boldsymbol\Delta'_{a_0}\in\mathbb{R}^{H\times d}$ denote the frozen parent's composite velocity from its generalist $\mathbf{v}'_G$ and specialist $\boldsymbol\Delta'_{a_0}$. We anchor the active policy's velocity along route $a_0$:
\begin{equation}
\mathcal{L}_{\mathrm{keep}}=\mathbb{E}_{c\sim\mathcal{R},\tau,\boldsymbol\epsilon}\Big\|
\mathbf{v}_G+\boldsymbol\Delta_{a_0}-\mathrm{sg}(\mathbf{v}')
\Big\|_F^2.
\label{eq:keep}
\end{equation}
When $\mathcal{I}_{\mathrm{new}}\neq\emptyset$, we additionally apply the same velocity anchor $\|\mathbf{v}_G+\boldsymbol\Delta_j-\mathrm{sg}(\mathbf{v}')\|_F^2$ to expanded experts $j\in\mathcal{I}_{\mathrm{new}}$ on $\mathcal{R}$, providing a safe fallback if a replay state is routed to a new branch.

\smallskip\noindent\textbf{Expert-facing representation anchoring.}
To keep retained experts compatible with the updating backbone, we anchor current prefix tokens $\mathbf{h}$ and layer-wise caches $(\mathbf{K},\mathbf{V})$ to the frozen parent features $(\mathbf{h}',\mathbf{K}',\mathbf{V}')$ on $\mathcal{R}$. Defining normalized squared Frobenius error $d(\mathbf{U},\mathbf{U}')=\|\mathbf{U}-\mathbf{U}'\|_F^2/\|\mathbf{U}'\|_F^2$ for any tensor pair $(\mathbf{U},\mathbf{U}')$, we minimize:
\begin{equation}
\mathcal{L}_{\mathrm{feature}}=\mathbb{E}_{c\sim\mathcal{R}}\frac{1}{2}\!\left[
 d(\mathbf{h},\mathbf{h}')+
 \frac{d(\mathbf{K},\mathbf{K}')+d(\mathbf{V},\mathbf{V}')}{2}
\right],
\label{eq:feat}
\end{equation}
which suppresses cross-attention drift on prior tasks while leaving $f_\phi$ adaptable to novel primitives outside the replay distribution.

\smallskip\noindent\textbf{Route retention and unified objective.}
Let $p_j=[\mathrm{softmax}(\tilde{\mathbf{z}})]_j$ be the routing probability of expert $j$ over $\tilde{\mathbf{z}}\in\mathbb{R}^{K+M_s}$. We penalize probability mass assigned to newly added experts on replay data via $\mathcal{L}_{\mathrm{ret}}=\mathbb{E}_{c\sim\mathcal{R}}\big(\sum_{j\in\mathcal{I}_{\mathrm{new}}}p_j\big)^2$ (which equals $0$ when $\mathcal{I}_{\mathrm{new}}=\emptyset$). With scalar weights $\lambda_{\mathrm{keep}},\lambda_{\mathrm{feature}},\lambda_{\mathrm{route}},\lambda_{\mathrm{ret}}>0$, the overall training objective for stage $s\ge2$ is:
\begin{equation}
\begin{aligned}
\mathcal{L}=\mathcal{L}_{\mathrm{act}}
&+\lambda_{\mathrm{keep}}\mathcal{L}_{\mathrm{keep}}
+\lambda_{\mathrm{feature}}\mathcal{L}_{\mathrm{feature}} \\
&+\lambda_{\mathrm{route}}\mathcal{L}_{\mathrm{route}}
+\lambda_{\mathrm{ret}}\mathcal{L}_{\mathrm{ret}}.
\end{aligned}
\label{eq:total}
\end{equation}

%% file: sec/4_experiments.tex
\section{Experiments}
\label{sec:exp}

Our experiments evaluate four core questions: (1)~whether primitive-level absorption and expansion sustain growing skill repertoires (\cref{sec:main}); (2)~how automatic kinematic novelty detection and observation-driven routing enable expert reuse in unseen compositions (\cref{sec:reuse}); (3)~whether \method scales to continual learning on the dual-arm robot G2A (\cref{sec:real}); and (4)~how chunk-routed residual experts and stage-wise anchoring retain backbone plasticity while mitigating interference (\cref{sec:plasticity}).

\subsection{Experimental setup and implementation}
\label{sec:setupexp}
\noindent\textbf{Continual-learning benchmarks.}
Following the official RoboCasa~\cite{nasiriany2026robocasa365} lifelong learning setting, we evaluate long-horizon continual adaptation across four stages (Stages~1--4, denoted S1--S4), where S1 consists of atomic skills and S2--S4 comprise multi-step compositions of S1 primitives.
Because RoboCasa composites mainly recombine S1 primitives, we also construct a continual benchmark from RoboTwin~2.0~\cite{chen2025robotwin2} to jointly test capacity expansion and compositional absorption across $50$ tasks over S1--S3: S1 covers $23$ atomic tasks (pick-and-place, pressing, articulation, reorientation, and bimanual lifting); S2 introduces $7$ novel-primitive tasks (shaking, tool contact, dragging, parallel grasping, and handover; $30$ cumulative); and S3 adds $16$ composite tasks combining earlier skills with unseen pouring and hanging ($46$ cumulative), plus $4$ held-out compositions for zero-shot evaluation.

\smallskip\noindent\textbf{Baselines and evaluation protocols.}
We compare \method against sequential fine-tuning (Seq.\ FT), weight interpolation (RETAIN~\cite{yadav2026retain}), and stage-isolated adapters routed via reconstruction error (CLARE~\cite{romer2026clare}). We further compare against matched-replay baselines trained under unified data, replay, and optimization settings: experience replay with velocity distillation inside a single policy (ER+KD), a parameter-matched wider Dense action head, and a layer-wise feedforward MoE baseline (FFN-MoE). After each stage, a single policy is evaluated across all learned tasks from $(o,\ell)$ over three seeds.

\smallskip\noindent\textbf{Implementation details.}
\label{sec:impl}
\method builds on $\pi_{0.5}$~\cite{intelligence2025pi05} with $K{=}4$ initial residual experts and chunk-level routing across all benchmarks. On RoboCasa, where S1 comprises atomic skills and S2--S4 comprise multi-step compositions, automatic kinematic novelty detection over S2--S4 classifies all sliding-window chunks within the four S1 clusters ($M_s{=}0$), triggering pure absorption into matched specialists and $g_\theta$ without adding experts. On RoboTwin, trajectories are segmented at gripper transitions, and silhouette-selected outlier clustering (\cref{eq:novelty-dist,eq:decision}) automatically determines expert growth (\cref{sec:reuse}): S2 adds two residual experts ($M_2{=}2$) while freezing all S1 experts ($\mathcal{I}_{\mathrm{abs}}{=}\emptyset$); S3 adds one expert ($M_3{=}1$) and unfreezes matched specialists ($\mathcal{I}_{\mathrm{abs}}$) together with $g_\theta$ to absorb familiar subtasks during composition. Continual stages use a replay ratio of $\rho{\approx}4.7\%$ on RoboTwin and $\rho{=}12.5\%$ on RoboCasa. Full training details are in the supplement.

\subsection{Main lifelong learning results}
\label{sec:main}

\begin{table}[!t]
\centering
\small
\setlength{\tabcolsep}{0pt}
\renewcommand{\arraystretch}{0.92}
\caption{\textbf{Cumulative success across learned tasks} (\%) on \mbox{RoboCasa} S2--S4 and \mbox{RoboTwin} S2.}
\label{tab:overall}
\begin{tabular*}{\columnwidth}{@{\extracolsep{\fill}}lcccc@{}}
\toprule
& \multicolumn{3}{c}{RoboCasa} & RoboTwin \\
\cmidrule(lr){2-4}\cmidrule(lr){5-5}
Method & S2 & S3 & S4 & S2 \\
\midrule
\multicolumn{5}{@{}l}{\emph{Continual-learning baselines}} \\
Seq.\ FT & 34.2 & 24.2 & 11.4 & 32.3 \\
RETAIN~\cite{yadav2026retain} & 48.7 & 39.8 & 27.7 & 49.5 \\
CLARE~\cite{romer2026clare} & 40.6 & 32.0 & 27.5 & 49.7 \\
\midrule
\multicolumn{5}{@{}l}{\emph{Matched-replay baselines}} \\
Dense & 43.1 & 34.9 & 29.0 & 58.5 \\
FFN-MoE & 42.3 & 34.2 & 29.9 & 51.0 \\
ER+KD & 55.8 & 47.6 & 39.5 & 57.0 \\
\midrule
\method & \textbf{58.1} & \textbf{48.4} & \textbf{41.4} & \textbf{63.5} \\
\bottomrule
\end{tabular*}
\end{table}

\begin{table*}[!t]
\centering
\small
\setlength{\tabcolsep}{0pt}
\renewcommand{\arraystretch}{0.92}
\caption{\textbf{Skill retention across prior cohorts and zero-shot composition} (success, \%). Top headers (S2--S4) denote evaluation stages; subcolumns (S1--S3) denote earlier task cohorts, plus Held-out zero-shot RoboTwin tasks.}
\label{tab:retention}
\begin{tabular*}{\textwidth}{@{\extracolsep{\fill}}lccccccccccc@{}}
\toprule
& & \multicolumn{6}{c}{RoboCasa} & \multicolumn{4}{c}{RoboTwin} \\
\cmidrule(lr){3-8}\cmidrule(lr){9-12}
& & S2 & \multicolumn{2}{c}{S3} & \multicolumn{3}{c}{S4} & S2 & \multicolumn{3}{c}{S3} \\
\cmidrule(lr){3-3}\cmidrule(lr){4-5}\cmidrule(lr){6-8}\cmidrule(lr){9-9}\cmidrule(lr){10-12}
Method & Replay & S1 & S1 & S2 & S1 & S2 & S3 & S1 & S1 & S2 & Held-out \\
\midrule
\multicolumn{12}{@{}l}{\emph{Continual-learning baselines}} \\
Seq.\ FT & $\times$ & 32.9 & 27.3 & 21.2 & 15.0 & 10.5 & 6.2 & 21.4 & 35.2 & 33.4 & 19.5 \\
RETAIN~\cite{yadav2026retain} & $\times$ & 54.1 & 52.5 & 28.8 & 42.8 & 22.2 & 8.7 & 52.5 & 59.8 & 40.9 & 2.5 \\
CLARE~\cite{romer2026clare} & $\times$ & 52.3 & 51.0 & 2.3 & 51.8 & 2.7 & 0.0 & 52.9 & 53.7 & 36.0 & 1.0 \\
\midrule
\multicolumn{12}{@{}l}{\emph{Matched-replay baselines}} \\
Dense & $\checkmark$ & 54.0 & 53.9 & 6.3 & 53.9 & 4.8 & 0.7 & 58.9 & 59.7 & 47.1 & 10.0 \\
FFN-MoE & $\checkmark$ & 54.0 & 54.3 & 2.7 & 57.1 & 1.2 & 0.0 & 52.5 & 44.2 & 30.9 & 5.0 \\
ER+KD & $\checkmark$ & 61.8 & 61.1 & \textbf{34.3} & 61.0 & 32.0 & 10.3 & 53.7 & 69.2 & 64.0 & 16.0 \\
\midrule
\method & $\checkmark$ & \textbf{64.8} & \textbf{64.3} & 32.0 & \textbf{63.5} & \textbf{33.8} & \textbf{12.7} & \textbf{61.5} & \textbf{70.4} & \textbf{79.4} & \textbf{22.0} \\
\bottomrule
\end{tabular*}
\end{table*}

\noindent\textbf{Cumulative performance across training stages.}
\Cref{tab:overall} reports cumulative success across all tasks seen up to each stage. On RoboCasa---where novelty detection absorbs S2--S4 into existing atomic specialists without adding experts---\method achieves the highest cumulative success in every continual stage ($58.1\%$ in S2, $48.4\%$ in S3, and $41.4\%$ in S4). On RoboTwin S2---where outlier clustering adds two experts for kinematically novel tasks---\method reaches $63.5\%$, outperforming ER+KD by $6.5\%$ and Seq.\ FT by $31.2\%$. Gains across both absorption-only and expansion-driven streams confirm that both branches of \cref{eq:decision} scale continual manipulation.

\smallskip\noindent\textbf{Mitigating interference on prior cohorts.}
\Cref{tab:retention} evaluates earlier task cohorts separately to measure skill retention. On RoboCasa, although ER+KD retains a slight edge on S2 tasks immediately after S3 ($34.3\%$ compared to $32.0\%$), by the final stage S4 \method achieves the highest retention across all prior cohorts ($63.5\%$ on S1, $33.8\%$ on S2, and $12.7\%$ on S3), improving over ER+KD by $2.5\%$, $1.8\%$, and $2.4\%$, whereas CLARE and FFN-MoE drop below $3.0\%$ on S2 and S3 composites. On RoboTwin, \method improves S1 retention after S2 by $7.8\%$ over ER+KD. After S3 compositions, \method retains $70.4\%$ on S1 and \textbf{79.4\%} on S2 tasks---outperforming ER+KD by \textbf{15.4\%} and CLARE by $43.4\%$---while also achieving the highest S2 new-task acquisition ($70.3\%$, $+2.9\%$ over ER+KD; supplement).

\subsection{Novelty detection, reuse, and routing analysis}
\label{sec:reuse}

\begin{figure*}[!t]
\centering
\includegraphics[width=0.92\textwidth]{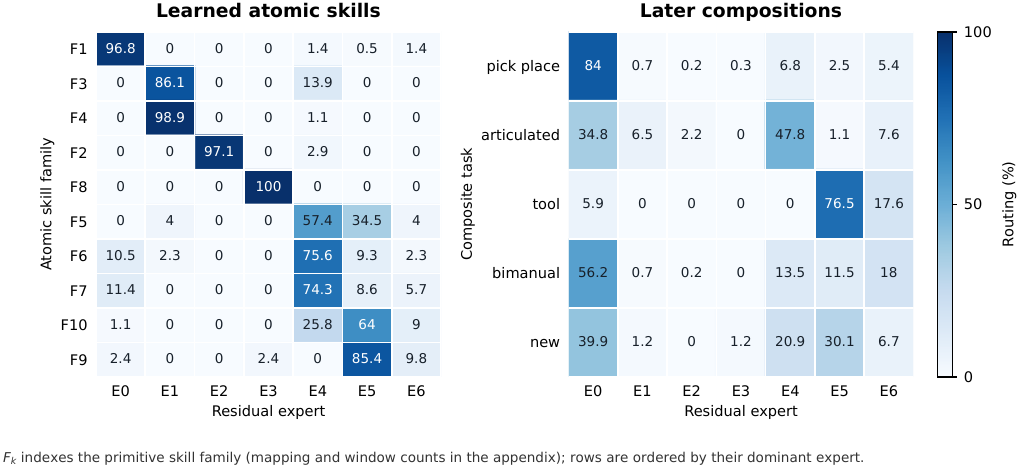}
\caption{\textbf{Primitive specialization persists and earlier experts are reused in later compositions.} Top-1 observation-driven routing (\%) of the S3 RoboTwin policy across $880$ atomic segments (left: S1 $F_1$--$F_4,F_8$; S2 $F_5$--$F_7,F_9$--$F_{10}$) and $2{,}088$ composite $50$-frame windows (right) over experts added in S1 ($E_0$--$E_3$), S2 ($E_4$--$E_5$), and S3 ($E_6$).}
\label{fig:primitive-reuse}
\end{figure*}

\noindent\textbf{Validation of kinematic novelty detection.}
Our distance threshold in \cref{eq:novelty-dist} reliably separates familiar and unseen primitives across all three domains. On RoboCasa S2--S4, $100\%$ of composite segments fall within the $95$th-percentile thresholds of the four S1 clusters ($M_2{=}M_3{=}M_4{=}0$), avoiding redundant expert growth across $60$ composite tasks. On our dual-arm robot G2A (\cref{sec:real}), clustering S2 trajectories identifies bimanual cloth folding and handover as a single outlier cluster beyond S1 single-arm thresholds ($M_2{=}1$). In S3, gripper segmentation splits container relocation into familiar right-arm placement ($72.5$ frames on average, matching the $69.5$-frame S1 cluster and absorbed) and unseen left-arm bin transport ($92.0$ frames, exceeding prior thresholds and adding one expert, $M_3{=}1$). On RoboTwin, outlier clustering likewise recovers two novel bimanual and tool-use clusters in S2 and one pouring-and-hanging cluster in S3, while error-driven competition routes $93.3\%$ of composite chunks back to inherited experts.

\smallskip\noindent\textbf{Zero-shot transfer to unseen compositions.}
Evaluated after S3 on the four unseen RoboTwin composites (\cref{tab:retention}), \method reaches \textbf{22.0\%} zero-shot success, surpassing ER+KD by $6.0\%$, whereas CLARE ($1.0\%$), RETAIN ($2.5\%$), and FFN-MoE ($5.0\%$) struggle to recombine skills learned across stages.

\smallskip\noindent\textbf{Routing verification of specialization and reuse.}
\Cref{fig:primitive-reuse} visualizes top-1 observation-driven routing of the S3 RoboTwin policy. S1 experts ($E_0$--$E_3$) align cleanly with atomic motion families ($96.8\%$ pick-and-place on $E_0$, $86.1\%$--$98.9\%$ pressing and articulation on $E_1$, $97.1\%$ reorientation on $E_2$, and $100\%$ bimanual lifting on $E_3$). S2 experts capture shaking, tool contact, and dragging on $E_4$ ($57.4\%$--$75.6\%$) as well as bimanual parallel grasping and handover on $E_5$ ($64.0\%$--$85.4\%$). In S3 composites, $84.0\%$ of pick-and-place windows route to $E_0$, $76.5\%$ of tool-use windows route to $E_5$, and articulated tasks combine approach and placement ($E_0$, $34.8\%$), handle dragging ($E_4$, $47.8\%$), and hinging ($E_1$, $6.5\%$). Even in composites containing unseen pouring and hanging, $93.3\%$ of chunks reuse inherited specialists ($39.9\%$ $E_0$, $30.1\%$ $E_5$, $20.9\%$ $E_4$) for shared grasping and transport, while the S3 expert $E_6$ accounts for $6.7\%$ of chunks in these composites.

\subsection{Real-robot lifelong learning}
\label{sec:real}
\noindent\textbf{Hardware curriculum and evaluation.}
We deploy \method on the dual-arm robot G2A with three $480{\times}640$ RGB cameras, predicting $24$-DoF actions ($22$ joint deltas and $2$ grippers) at $30$\,Hz in $50$-step chunks. We collect $450$ demonstrations across five tasks over S1--S3: S1 trains on right-arm pick-and-place and washing-machine door articulation ($200$ demos); S2 adds bimanual cloth folding and garment handover ($160$ demos); and S3 introduces two-phase container relocation ($90$ demos) that places a plush toy into a bin with the right arm before grasping the bin handle with the left arm and transporting it to a target zone. Novelty detection initializes four experts in S1, adds one expert for the bimanual cluster in S2, and absorbs right-arm placement while adding one expert for left-arm bin transport in S3. All methods use unified training settings with a $12.5\%$ replay ratio over $25$ physical trials per task after each stage.

\begin{table}[!t]
\centering
\small
\setlength{\tabcolsep}{0pt}
\renewcommand{\arraystretch}{0.92}
\caption{\textbf{Real-robot lifelong learning on G2A} (success, \%).}
\label{tab:realrobot}
\begin{tabular*}{\columnwidth}{@{\extracolsep{\fill}}llccc@{}}
\toprule
Training stage & Evaluated tasks & ER+KD & FFN-MoE & \method \\
\midrule
S1 & S1 & 46.0 & 14.0 & \textbf{60.0} \\
\midrule
S2 & S1 & 36.0 & 10.0 & \textbf{52.0} \\
   & All learned & 34.0 & 8.0 & \textbf{49.0} \\
\midrule
S3 & S1 & 26.0 & 8.0 & \textbf{44.0} \\
   & S2 & 4.0 & 0.0 & \textbf{22.0} \\
   & All learned & 28.0 & 5.6 & \textbf{44.8} \\
\bottomrule
\end{tabular*}
\end{table}

\smallskip\noindent\textbf{Physical retention and acquisition results.}
As shown in \cref{tab:realrobot}, \method outperforms ER+KD and FFN-MoE across every evaluated cohort in S1--S3. Separating pick-and-place and articulation across residual experts raises initial S1 acquisition to \textbf{60.0\%}, exceeding single-head ER+KD ($46.0\%$) by $14.0\%$. From S1 to S3, \method drops by $16.0\%$ ($60.0\%\to\mathbf{44.0\%}$) compared with a $20.0\%$ drop for ER+KD ($46.0\%\to 26.0\%$), combining stronger initial learning with $4.0\%$ less forgetting for a $+18.0\%$ final S1 margin. On S2 bimanual tasks, \method reaches $46.0\%$ initial acquisition, exceeding ER+KD ($32.0\%$) by $14.0\%$, and retains \textbf{22.0\%} after S3 (a $24.0\%$ drop) whereas ER+KD nearly collapses to $4.0\%$ (a $28.0\%$ drop), yielding a $+18.0\%$ retention margin. On S3 composition, \method achieves \textbf{92.0\%} acquisition ($+12.0\%$ over ER+KD), reaching \textbf{44.8\%} overall success across all five tasks---surpassing ER+KD by \textbf{16.8\%} and FFN-MoE by $39.2\%$.

\subsection{Ablation studies}
\label{sec:plasticity}
\noindent\textbf{Chunk-routed residual experts.}
Comparing parameter-matched Dense and FFN-MoE baselines under unified replay settings confirms that modular structure---rather than parameter count---drives retention (\cref{tab:retention}). After RoboTwin S3, Dense retains $59.7\%$ on S1 and $47.1\%$ on S2, while FFN-MoE drops to $44.2\%$ on S1 and $30.9\%$ on S2, trailing \method ($70.4\%$ on S1 and $79.4\%$ on S2) by \textbf{32.3\% and 48.5\%} on S2 primitives. By RoboCasa S4, both Dense and FFN-MoE fall below $4.8\%$ on S2 whereas \method maintains $33.8\%$, showing that chunk-level residual routing preserves intra-chunk temporal coherence.

\begin{table}[!t]
\centering
\small
\setlength{\tabcolsep}{0pt}
\renewcommand{\arraystretch}{0.92}
\caption{\textbf{S3 allocation ablation on RoboTwin} (success, \%).}
\label{tab:allocation}
\begin{tabular*}{\columnwidth}{@{\extracolsep{\fill}}lccc@{}}
\toprule
& \multicolumn{2}{c}{Prior-cohort retention} & Zero-shot \\
\cmidrule(lr){2-3}\cmidrule(lr){4-4}
Allocation rule & S1 & S2 & Held-out \\
\midrule
Uniform Absorb & 65.6 & 69.1 & 17.0 \\
Uniform Expand & 65.4 & 66.0 & 17.5 \\
\method (Absorb + Expand) & \textbf{70.4} & \textbf{79.4} & \textbf{22.0} \\
\bottomrule
\end{tabular*}
\end{table}

\smallskip\noindent\textbf{Primitive-aware allocation.}
To isolate \cref{eq:decision} under matched training steps and replay, \cref{tab:allocation} branches RoboTwin S3 from the same S2 checkpoint against Uniform Absorb (updating all six existing experts) and parameter-matched Uniform Expand (appending one stage-level expert without primitive separation). Uniform Absorb lowers S2 retention by $10.3\%$ (to $69.1\%$), while Uniform Expand fails to consolidate shared subtasks into inherited specialists and loses $13.4\%$ on S2 ($66.0\%$). Selective absorption and expansion improves S1 retention by up to $5.0\%$, S2 retention by \textbf{13.4\%}, and zero-shot composition by up to $5.0\%$ (\textbf{22.0\%}).

\begin{table}[!t]
\centering
\small
\setlength{\tabcolsep}{0pt}
\renewcommand{\arraystretch}{0.92}
\caption{\textbf{S2 backbone and anchoring ablation} (success, \%).}
\label{tab:plasticity}
\begin{tabular*}{\columnwidth}{@{\extracolsep{\fill}}lccccc@{}}
\toprule
& \multicolumn{2}{c}{Trainable} & & \multicolumn{2}{c}{Stage 2} \\
\cmidrule(lr){2-3}\cmidrule(lr){5-6}
Variant & Backbone & Gen.\ act. & $\mathcal{L}_{\mathrm{feature}}$ & Old & New \\
\midrule
\multicolumn{6}{@{}l}{\emph{RoboTwin}} \\
New experts only & & & & 60.0 & 47.9 \\
\quad + double steps & & & & 58.5 & 45.0 \\
\quad + action branch & & $\checkmark$ & & 51.0 & 48.6 \\
\method\ w/o $\mathcal{L}_{\mathrm{feature}}$ & $\checkmark$ & & & 61.0 & 65.0 \\
\method & $\checkmark$ & & $\checkmark$ & \textbf{62.0} & \textbf{69.3} \\
\midrule
\multicolumn{6}{@{}l}{\emph{RoboCasa}} \\
New expert only & & & & 50.5 & 8.0 \\
\method & $\checkmark$ & $\checkmark$ & $\checkmark$ & \textbf{57.0} & \textbf{37.8} \\
\bottomrule
\end{tabular*}
\end{table}

\smallskip\noindent\textbf{Adaptable backbone and stage-wise anchoring.}
\Cref{tab:plasticity} shows why freezing $f_\phi$ bottlenecks continual learning and isolates representation anchoring under the fast ablation protocol. With $f_\phi$ frozen and only new experts trained, S2 new-task acquisition stalls at $47.9\%$ on RoboTwin and collapses to \textbf{8.0\%} on RoboCasa; neither doubling training steps ($45.0\%$) nor unfreezing the generalist action branch ($48.6\%$) resolves this perceptual bottleneck. Unfreezing $f_\phi$ drives the main plasticity gain---already raising RoboTwin new-task acquisition to $65.0\%$ ($+17.1\%$) with velocity anchoring alone (\method\ w/o $\mathcal{L}_{\mathrm{feature}}$)---and reaches \textbf{69.3\%} on RoboTwin ($+21.4\%$ over frozen $f_\phi$) and \textbf{37.8\%} on RoboCasa ($+29.8\%$) under full \method, alongside $62.0\%$ and $57.0\%$ old-task retention. Comparing \method against \method\ w/o $\mathcal{L}_{\mathrm{feature}}$ isolates feature anchoring: removing $\mathcal{L}_{\mathrm{feature}}$ increases normalized prefix-token drift by $50\times$ and KV-cache drift by $47\times$, whereas adding $\mathcal{L}_{\mathrm{feature}}$ improves old-task retention by $1.0\%$ ($61.0\%\to 62.0\%$), new-task acquisition by $4.3\%$ ($65.0\%\to 69.3\%$), and overall S2 success from $62.6\%$ to $65.0\%$. In a separate expanded-expert fallback ablation in the supplement, removing replay velocity anchoring on newly added experts reduces prior-stage retention by $11.0\%$ relative to the \emph{New experts only} row ($60.0\%\to 49.0\%$).

%% file: sec/5_conclusion.tex
\section{Conclusion}
\label{sec:conclusion}

We presented \textbf{\method}, a primitive-aware residual mixture-of-experts policy for continual robot manipulation.
Using kinematic clustering and self-supervised error competition, \method \emph{absorbs} familiar primitives into existing action experts and \emph{expands} capacity guided by kinematic novelty.
To keep inherited specialists functional as the shared vision-language backbone adapts to new tasks, stage-wise feature and velocity anchoring stabilizes expert-facing representations and output velocities while mitigating interference on earlier skills.
Experiments on the official RoboCasa lifelong benchmark, our continual RoboTwin~2.0 benchmark, and five tasks on the dual-arm robot G2A show clear improvements in multi-stage skill retention, new-task acquisition, and zero-shot composition.
These findings highlight primitive-level adaptation as a practical path toward open-ended continual manipulation.

%% file: sec/6_llm.tex
\section*{LLM Usage Statement}

ChatGPT, Claude, and Gemini were used as assistive tools with moderate involvement: (i) language editing and polishing, (ii) literature search and related-work organization, (iii) code implementation, debugging, and data analysis, and (iv) plotting and table formatting. The authors developed the research ideas, designed the method, made all experimental decisions, interpreted the results, and verified the technical content. No images, diagrams, or videos were generated by AI tools. The authors take full responsibility for the content of this paper.

%% file: sec/X_suppl.tex
\clearpage
\maketitlesupplementary

\section{Stage-wise new-task acquisition}
\label{sec:supp-acquisition}

\Cref{tab:acquisition} reports success rates on newly introduced task cohorts immediately after each training stage, covering $7$ and $16$ tasks in RoboTwin Stages~2--3 (S2--S3) and $20$ composite tasks per stage in RoboCasa S2--S4, building on $65$ atomic S1 tasks, to complement \cref{tab:retention}. Whereas continual-learning baselines without replay (RETAIN, CLARE) and unstructured capacity variants (Dense, FFN-MoE) suffer severe plasticity loss in later stages, dropping to $3.6\%$--$30.8\%$ on RoboTwin S3 and $0.2\%$--$9.3\%$ on RoboCasa S3--S4, \method achieves the highest new-task acquisition in every column. On RoboTwin S2 it reaches $70.3\%$, outperforming ER+KD by $2.9$ percentage points and Seq.\ FT by $2.3$ percentage points, while on RoboTwin S3 and RoboCasa S2--S4 it attains $50.9\%$, $39.3\%$, $18.3\%$, and $8.8\%$, respectively. Aggregating all $46$ learned RoboTwin tasks after S3 as the task-count-weighted average of the S1, S2, and S3 cohorts (\cref{tab:acquisition}, ``All $46$''), \method reaches $65.0\%$ cumulative success, exceeding ER+KD ($59.9\%$) by $5.1$ percentage points. Seq.\ FT and single-head ER+KD fit current-stage tasks at the expense of severe forgetting or cross-stage interference on earlier skills, whereas \method maintains higher cumulative success and final-stage skill retention (\cref{tab:overall,tab:retention}).

\begin{table*}[t]
\centering
\small
\setlength{\tabcolsep}{0pt}
\renewcommand{\arraystretch}{1.12}
\caption{\textbf{New-task acquisition across training stages} (success, \%). Columns report performance on the newly introduced task cohort at the end of each stage, averaged over three evaluation seeds under the full evaluation protocol. The ``All $46$'' column reports cumulative success over all learned RoboTwin tasks after S3, computed as the task-count-weighted average ($23{:}7{:}16$) of the S1 and S2 cohort results in \cref{tab:retention} and the S3 cohort in this table.}
\label{tab:acquisition}
\begin{tabular*}{\textwidth}{@{\extracolsep{\fill}}lccccccc@{}}
\toprule
& & \multicolumn{3}{c}{RoboTwin} & \multicolumn{3}{c}{RoboCasa} \\
\cmidrule(lr){3-5}\cmidrule(lr){6-8}
Method & Replay & S2 & S3 & All $46$ & S2 & S3 & S4 \\
\midrule
\multicolumn{8}{@{}l}{\emph{Continual-learning baselines}} \\
Seq.\ FT & $\times$ & 68.0 & 45.4 & 38.5 & 38.5 & 17.2 & 5.7 \\
RETAIN~\cite{yadav2026retain} & $\times$ & 39.7 & 3.6 & 37.4 & 31.0 & 9.3 & 3.2 \\
CLARE~\cite{romer2026clare} & $\times$ & 39.4 & 9.4 & 35.6 & 2.3 & 0.2 & 0.5 \\
\midrule
\multicolumn{8}{@{}l}{\emph{Matched-replay baselines}} \\
Dense & $\checkmark$ & 57.1 & 30.8 & 47.7 & 7.5 & 1.7 & 0.7 \\
FFN-MoE & $\checkmark$ & 46.0 & 18.1 & 33.1 & 4.2 & 0.2 & 0.5 \\
ER+KD & $\checkmark$ & 67.4 & 44.6 & 59.9 & 36.5 & 17.2 & 6.3 \\
\midrule
\method & $\checkmark$ & \textbf{70.3} & \textbf{50.9} & \textbf{65.0} & \textbf{39.3} & \textbf{18.3} & \textbf{8.8} \\
\bottomrule
\end{tabular*}
\end{table*}

\section{Complete real-robot evaluation}
\label{sec:supp-realrobot}

\Cref{tab:realrobot-detailed} provides the complete stage-by-stage breakdown on the dual-arm G2A hardware setup across $2$ atomic tasks in S1, $2$ bimanual tasks in S2, and $1$ two-phase compositional task in S3, evaluated over $25$ physical trials per task at each stage. \Cref{fig:realrobot-tasks} illustrates all five tasks. In S3, demonstrations execute a two-phase sequence averaging $385.4$ frames at $30$\,Hz, up from $177.5$ frames in S1 pick-and-place. The right arm first places the plush toy into the green storage box ($72.5$ frames on average), matching the $69.5$-frame S1 primitive for absorption into the matched S1 specialist; the contralateral left arm then grasps the box handle and transports the container into the marked blue boundary zone ($92.0$ frames), exceeding prior cluster thresholds and adding one expanded expert. Here $72.5$ and $92.0$ denote the gripper-open--close segmented contact and transport segments; the remaining frames of each episode cover approach, retraction, and inter-phase transit, which are likewise chunked and routed as ordinary $50$-frame windows. Across all three stages, \method outperforms ER+KD and FFN-MoE on every cohort. It establishes a stronger initial skill library in S1 ($60.0\%$, exceeding ER+KD at $46.0\%$ by $14.0\%$) with a smaller drop from S1 to S3 ($16.0\%$ compared with a $20.0\%$ drop for ER+KD). Furthermore, \method achieves higher new-task acquisition in both S2 ($46.0\%$, $+14.0\%$ over ER+KD) and S3 ($92.0\%$, $+12.0\%$ over ER+KD), while mitigating the severe forgetting suffered by ER+KD on bimanual S2 skills ($22.0\%$ retained after S3 with a $24.0\%$ drop, whereas ER+KD drops by $28.0\%$ to $4.0\%$).

\paragraph{Real-robot inference latency.}
At deployment on the G2A setup with a single NVIDIA RTX~4090, predicting a $50$-step ($1.67$\,s at $30$\,Hz) action chunk from three $480{\times}640$ RGB views and language instructions takes $103$\,ms for single-branch ER+KD, $116$\,ms for \method, and $176$\,ms for FFN-MoE (\cref{tab:realrobot-detailed}, bottom row). Because \method selects a single residual specialist $S_{k^{\star}}$ once per chunk prior to flow integration and shares the VLM prefix tokens $h$ and key-value caches $(K, V)$ with the generalist branch $G$, it adds only $13$\,ms over the single-head baseline while keeping inference latency constant as the expert library grows across stages ($K{=}4\to 6$). In contrast, FFN-MoE incurs a $73$\,ms overhead due to layer-wise token-level top-$2$ routing across transformer blocks during every Euler integration step, yet still trails \method by $39.2\%$ in final-stage success. All three policies run well within the $1.67$\,s chunk execution horizon, supporting smooth $30$\,Hz real-time control.

\begin{table}[t]
\centering
\small
\setlength{\tabcolsep}{0pt}
\renewcommand{\arraystretch}{1.12}
\caption{\textbf{Complete stage-wise real-robot evaluation} (success, \%) and \textbf{chunk inference latency} (ms, measured on a single NVIDIA RTX~4090) on G2A. Earlier-stage rows measure retention, current-stage rows measure new-task acquisition, and ``All learned'' aggregates all tasks encountered up to that training stage over $25$ trials per task.}
\label{tab:realrobot-detailed}
\begin{tabular*}{\columnwidth}{@{\extracolsep{\fill}}llccc@{}}
\toprule
Training stage & Evaluated tasks & ER+KD & FFN-MoE & \method \\
\midrule
S1 & S1 & 46.0 & 14.0 & \textbf{60.0} \\
\midrule
S2 & S1 & 36.0 & 10.0 & \textbf{52.0} \\
   & S2 & 32.0 & 6.0 & \textbf{46.0} \\
   & All learned & 34.0 & 8.0 & \textbf{49.0} \\
\midrule
S3 & S1 & 26.0 & 8.0 & \textbf{44.0} \\
   & S2 & 4.0 & 0.0 & \textbf{22.0} \\
   & S3 & 80.0 & 12.0 & \textbf{92.0} \\
   & All learned & 28.0 & 5.6 & \textbf{44.8} \\
\midrule
\multicolumn{2}{@{}l}{Chunk latency (ms) $\downarrow$} & \textbf{103} & 176 & 116 \\
\bottomrule
\end{tabular*}
\end{table}

\begin{figure*}[t]
\centering
\def\shot#1{\raisebox{-0.5\height}{\includegraphics[width=0.185\textwidth]{figures/realrobot/#1.jpg}}}
\setlength{\tabcolsep}{3pt}
\renewcommand{\arraystretch}{1.1}
\begin{tabular}{@{}lcccc@{}}
\toprule
S1 pick-and-place & \shot{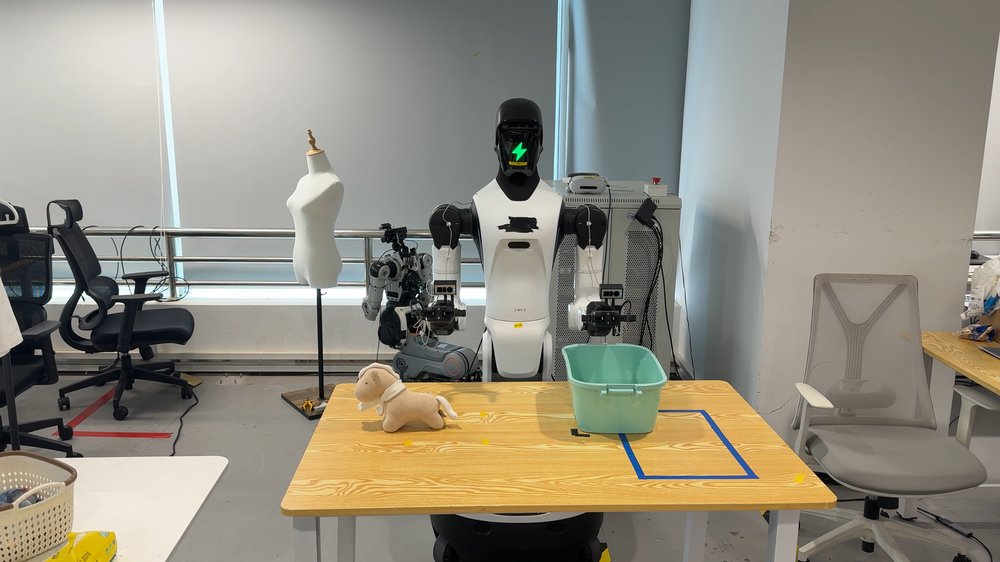} & \shot{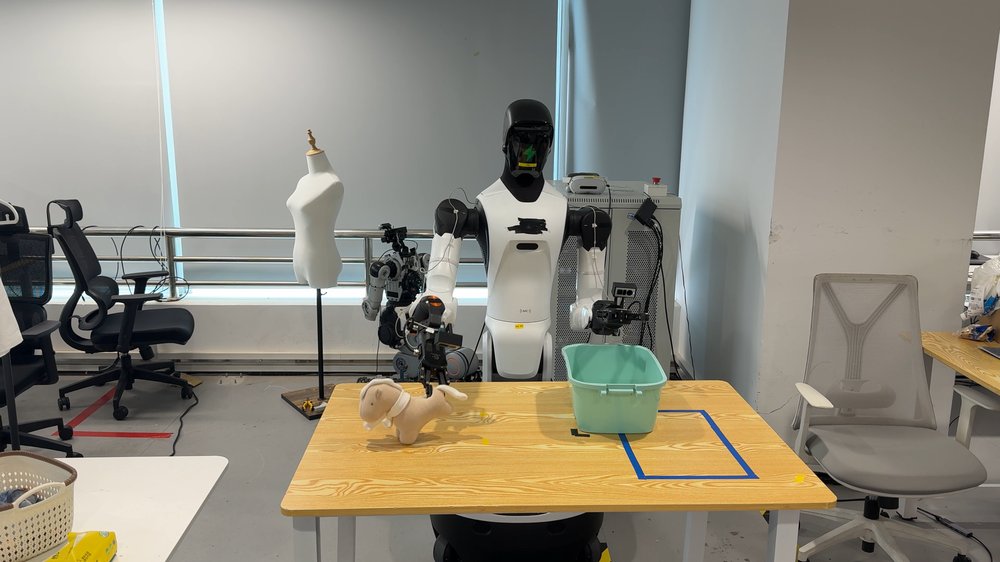} & \shot{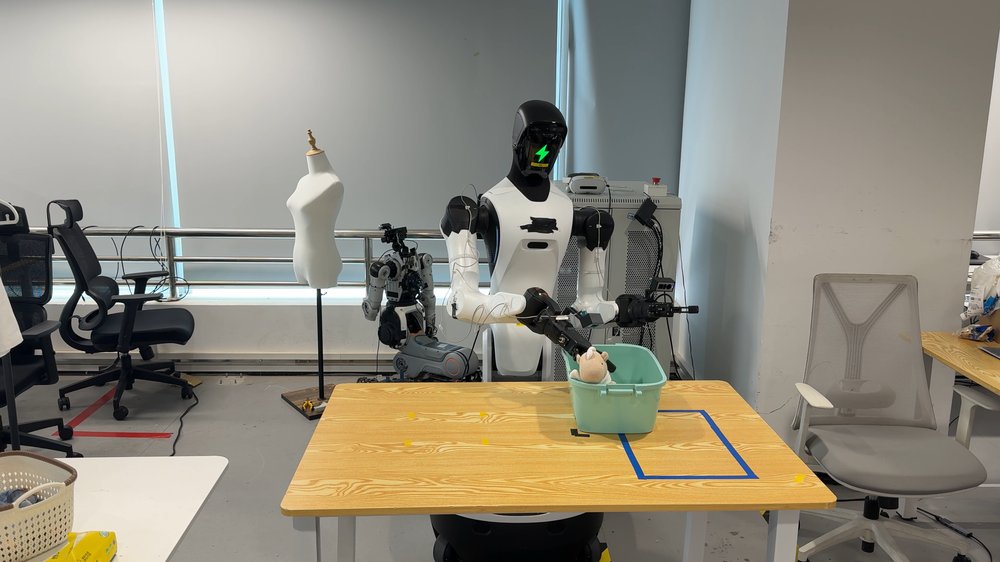} &  \\
\addlinespace[2pt]
S1 articulation & \shot{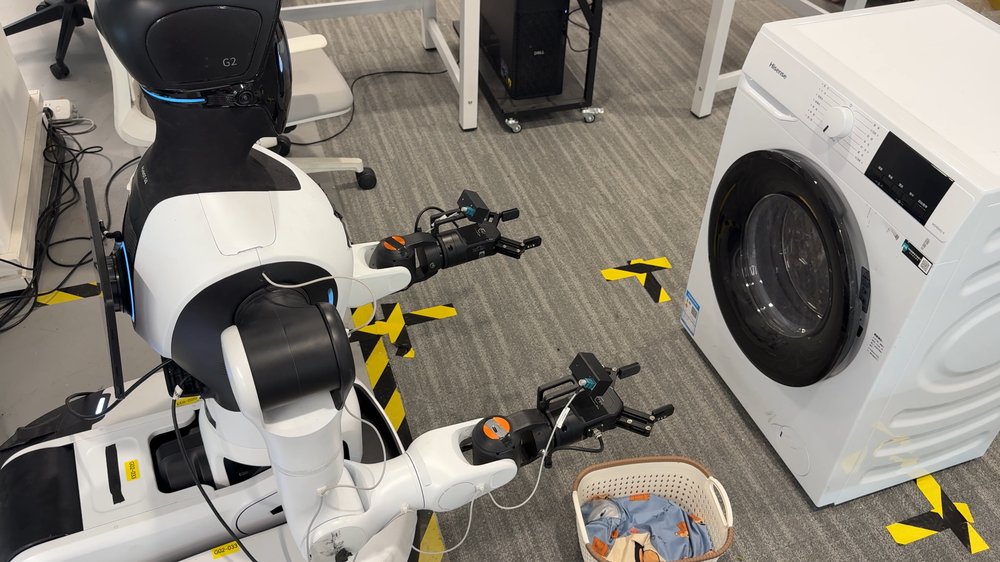} & \shot{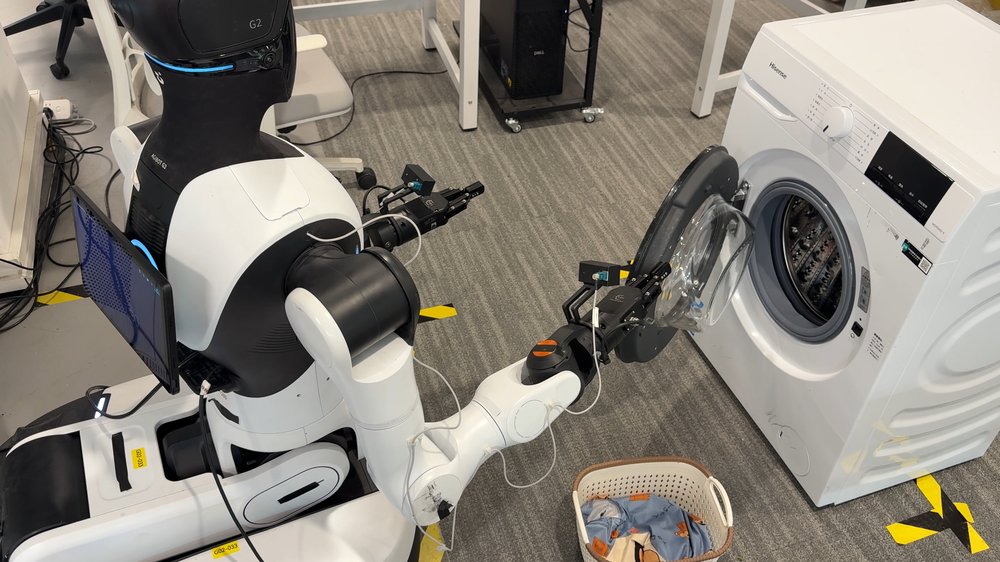} & \shot{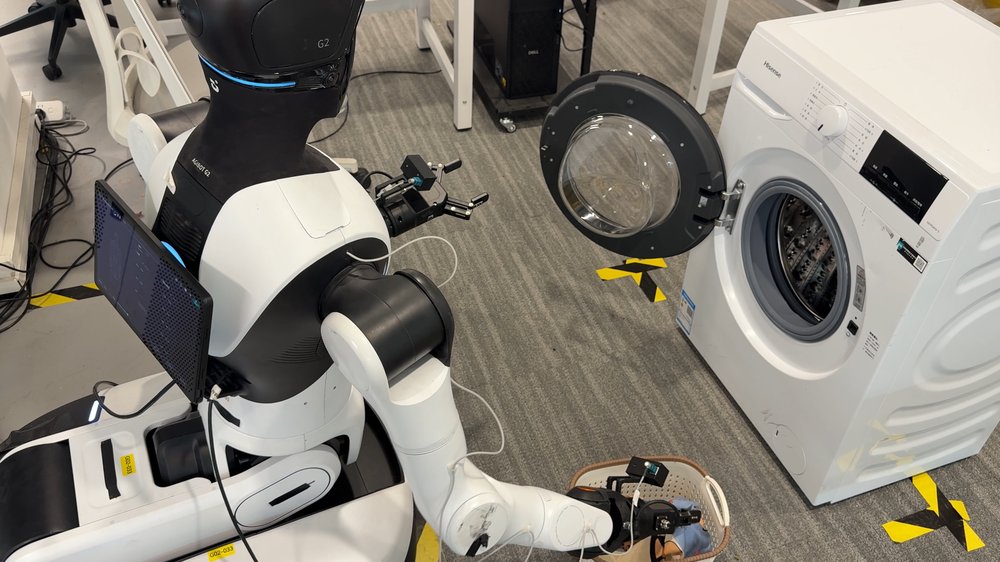} & \shot{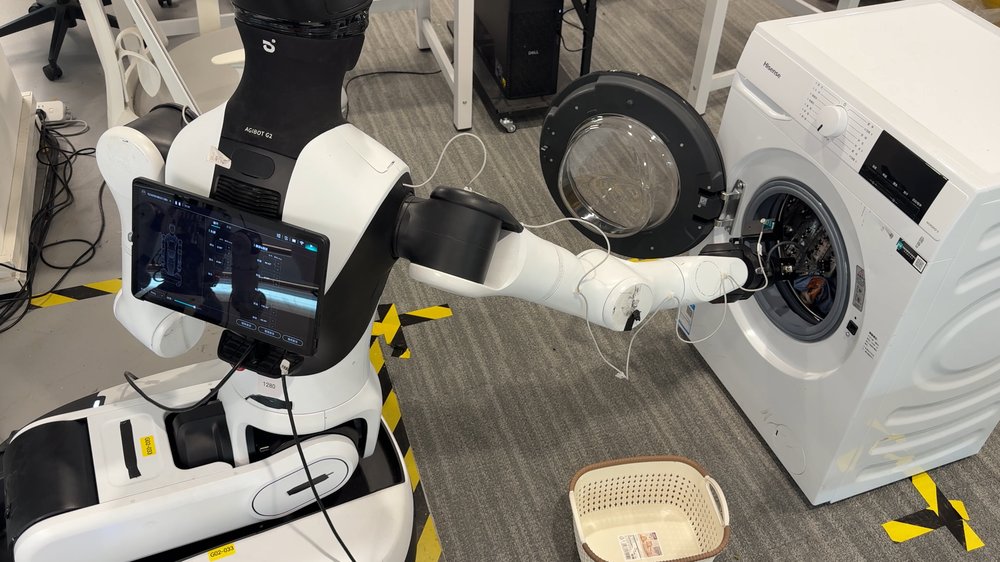} \\
\addlinespace[2pt]
S2 cloth folding & \shot{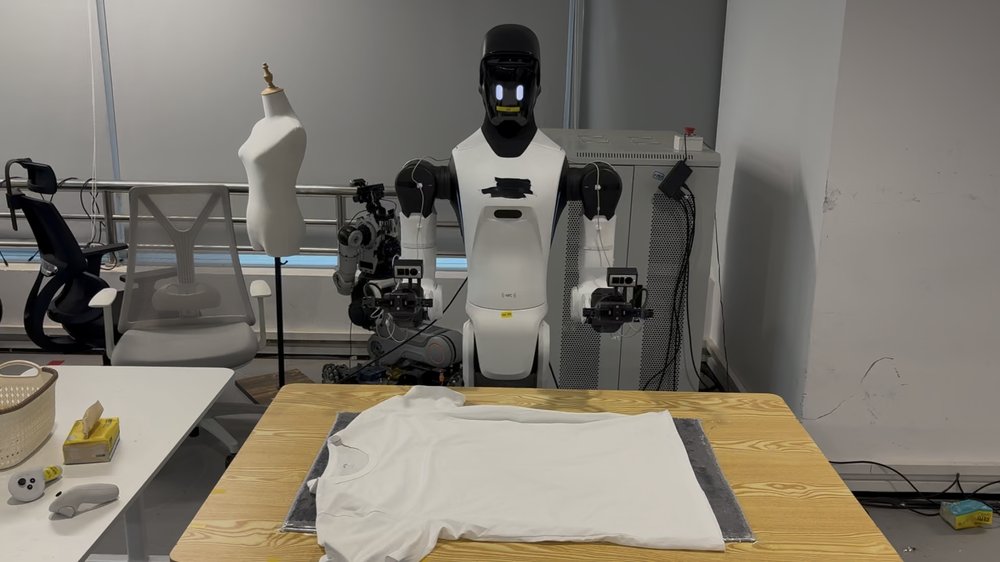} & \shot{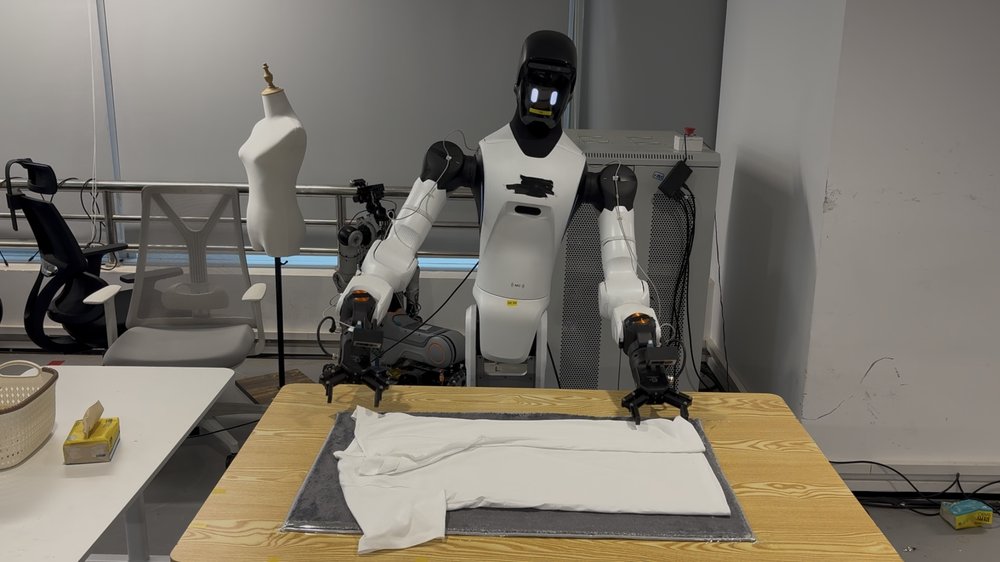} & \shot{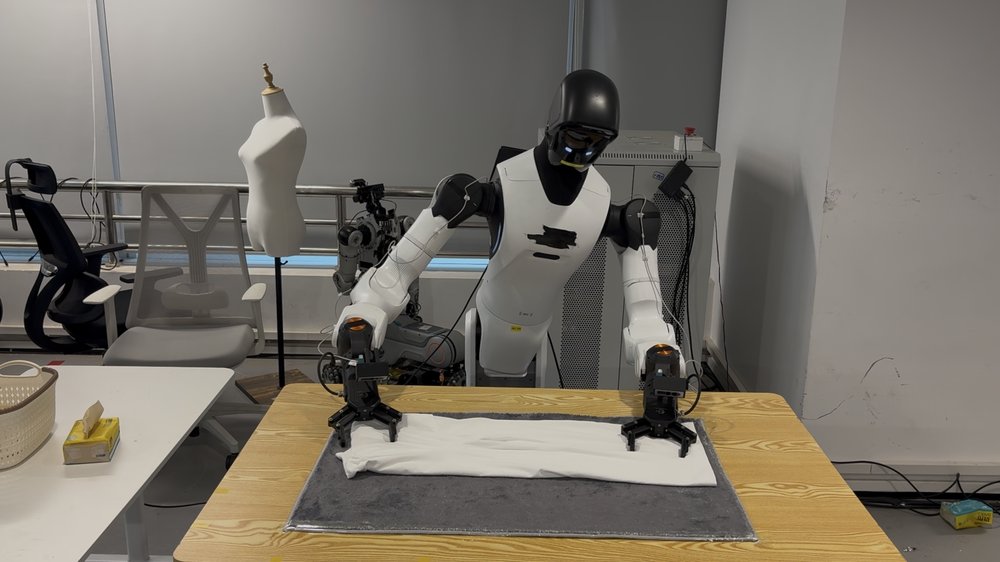} & \shot{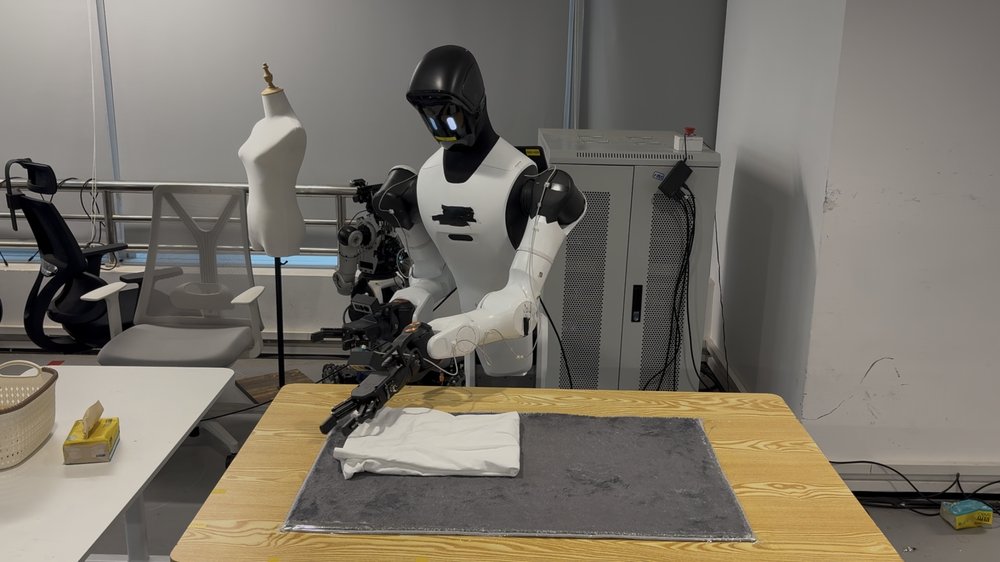} \\
\addlinespace[2pt]
S2 handover & \shot{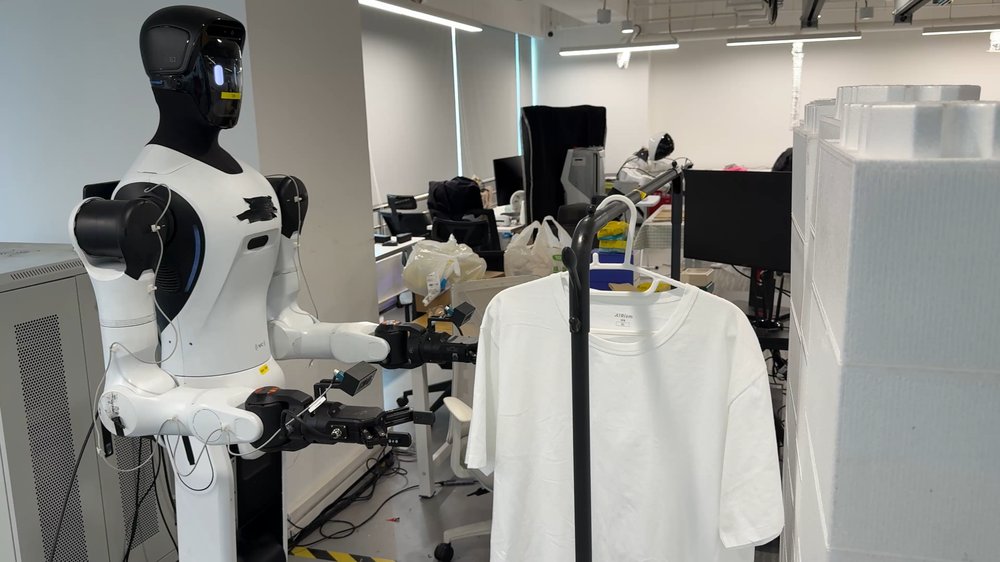} & \shot{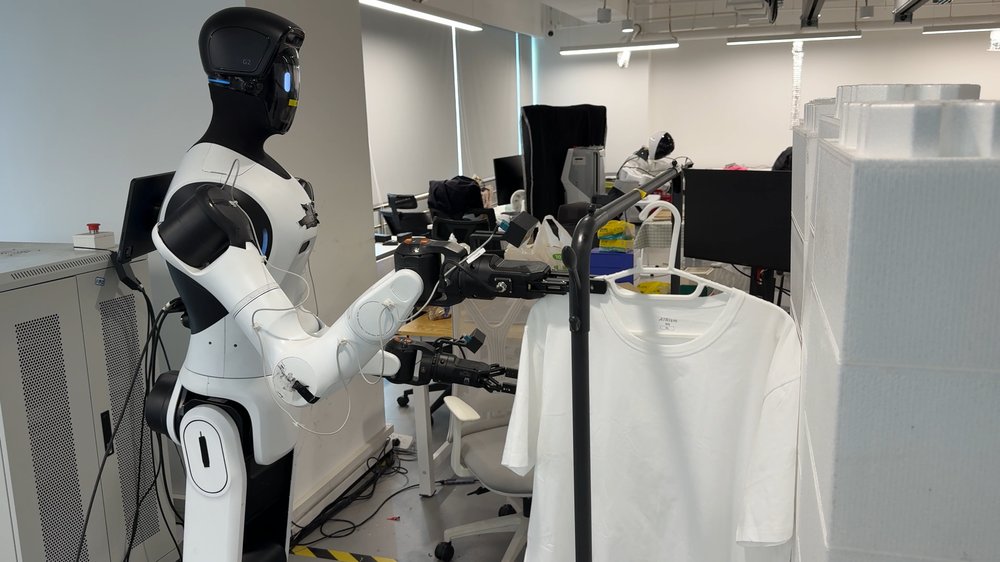} & \shot{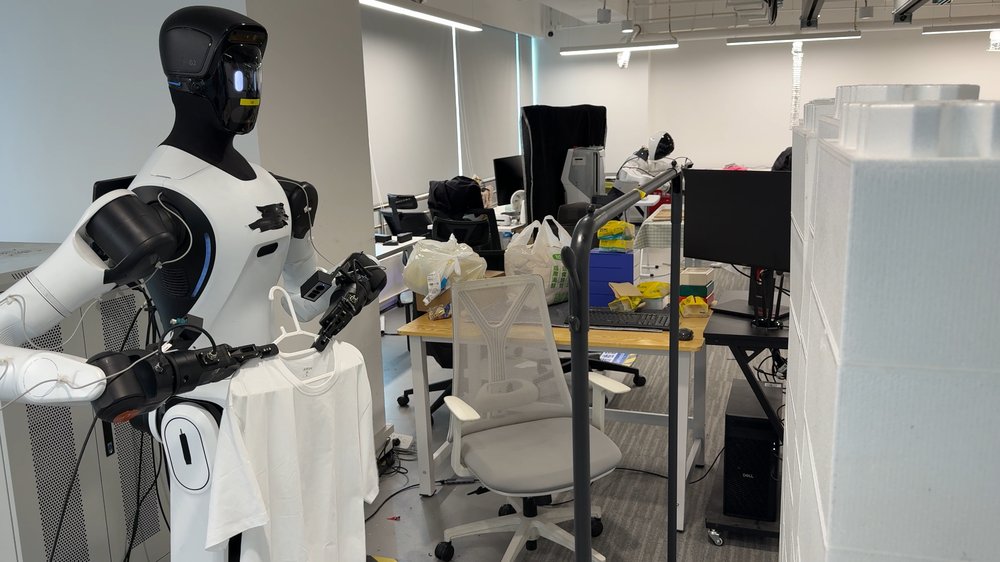} & \shot{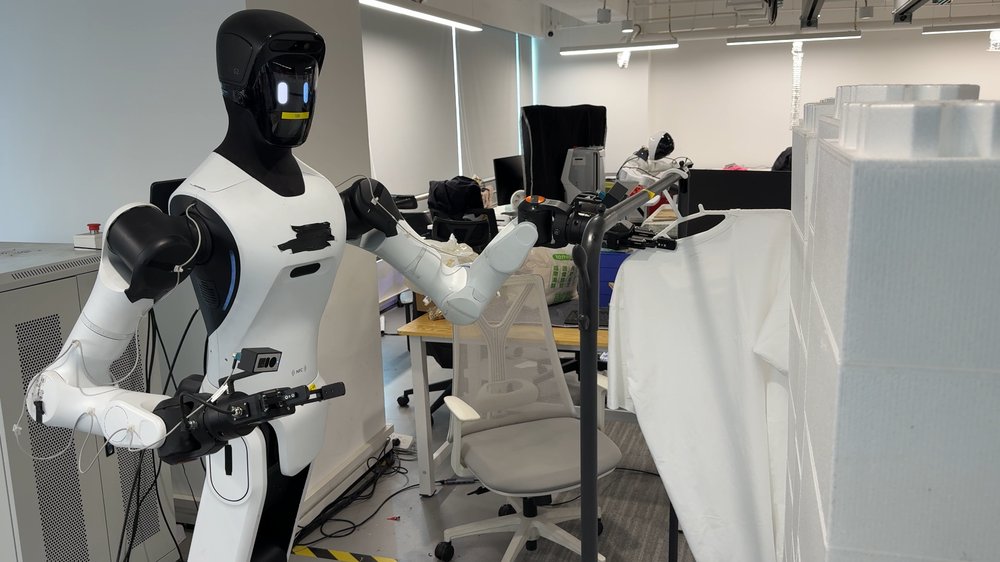} \\
\addlinespace[2pt]
S3 container relocation & \shot{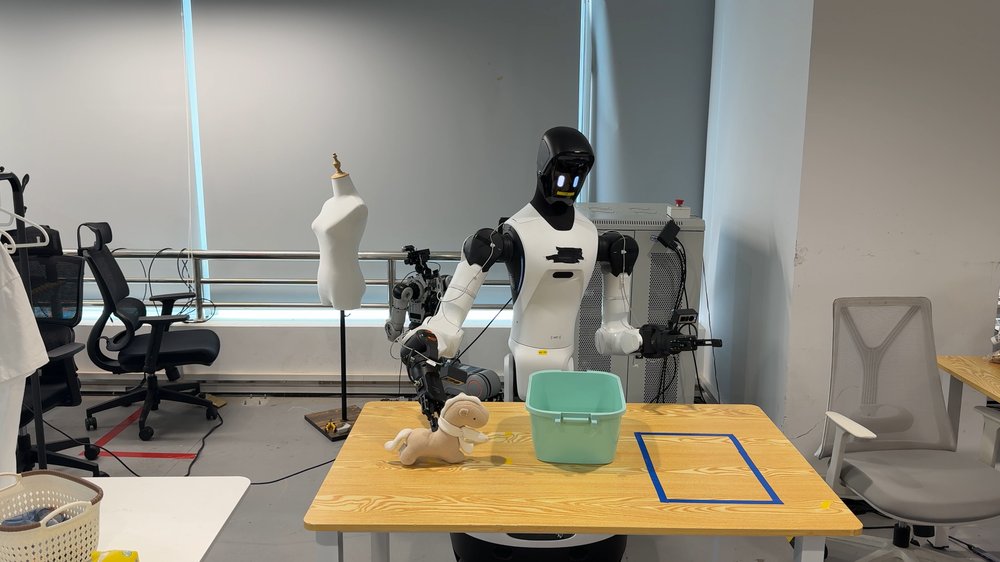} & \shot{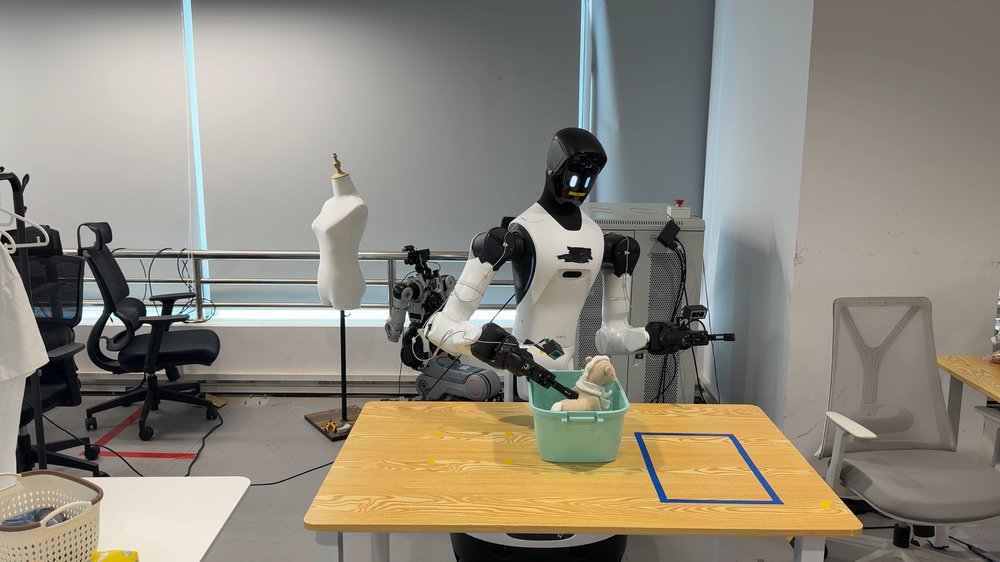} & \shot{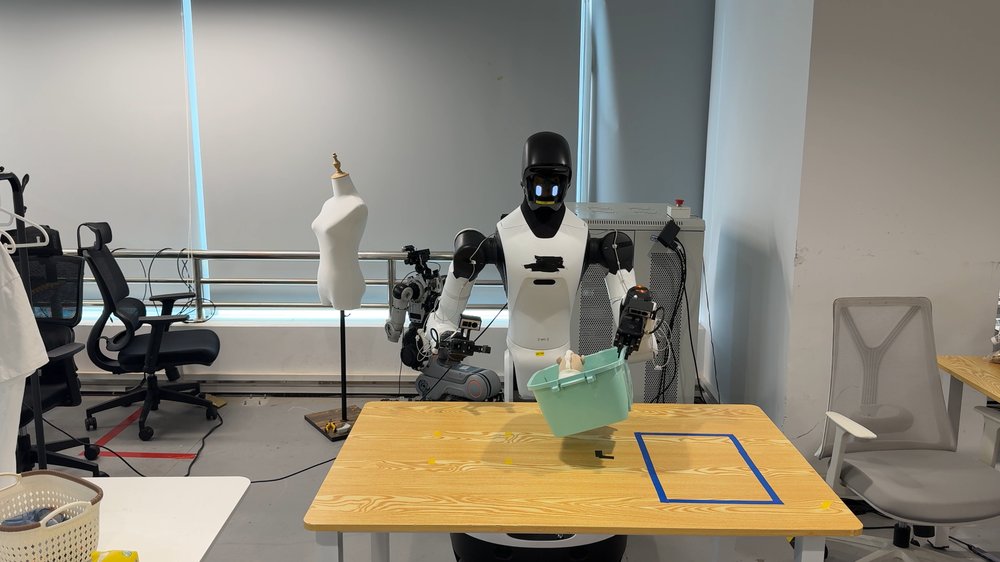} & \shot{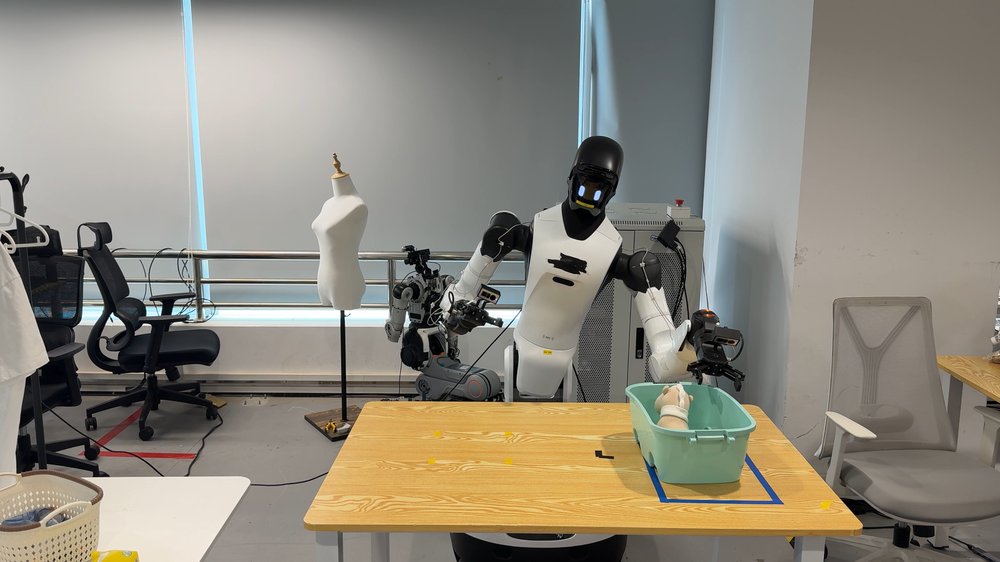} \\
\bottomrule
\end{tabular}
\caption{\textbf{Real-robot task suite on the dual-arm G2A platform.} Each row shows three to four temporal keyframes captured under three $480{\times}640$ RGB cameras with $24$-DoF action control. S1 trains on right-arm pick-and-place of a plush toy into a green storage bin and articulation of a washing-machine door; S2 adds bimanual cloth folding and bimanual garment handover on a rack; S3 introduces two-phase container relocation combining familiar right-arm toy-into-bin placement (absorbed) with contralateral left-arm transport of the green bin into a blue goal region (expanded).}
\label{fig:realrobot-tasks}
\end{figure*}

\section{Extended architectural and training analyses}
\label{sec:supp-analyses}

We present additional design-space ablations on uniform S3 allocation controls and anchoring configurations.

\subsection{Protocol details for uniform S3 allocation controls}
\label{sec:supp-absorbexpand}

In \cref{tab:allocation} of the main paper, \emph{Uniform Absorb}, \emph{Uniform Expand}, and \method are trained under matched compute and replay settings on RoboTwin S3. All three variants branch from the identical S2 checkpoint ($6$ existing residual experts) for $30$k optimization steps using batch size $128$, replay ratio $\rho{\approx}4.7\%$, identical backbone and generalist unfreezing schedules, and matched anchoring weights ($\lambda_{\mathrm{keep}}{=}\lambda_{\mathrm{feature}}{=}1.0$). \emph{Uniform Absorb} ($M_3{=}0$, $6$ experts) updates all six existing experts without instantiating new expert parameters ($\mathcal{I}_{\mathrm{abs}}{=}\{1,\ldots,6\}$), routing all composite chunks across the six inherited specialists. \emph{Uniform Expand} ($M_3{=}1$, $7$ experts, parameter-matched to \method) freezes prior experts and appends a single stage-level residual expert and router column without primitive-level separation. Both uniform controls trail \method ($70.4\%$ on S1, $79.4\%$ on S2, and $22.0\%$ zero-shot) across all metrics: S1 retention drops to $65.4\%$--$65.6\%$ (trailing by $4.8$--$5.0$ percentage points), S2 retention drops to $66.0\%$--$69.1\%$ (trailing by $10.3$--$13.4$ percentage points), and zero-shot composition drops to $17.0\%$--$17.5\%$ (trailing by $4.5\%$--$5.0\%$), confirming the benefit of primitive-selective absorption and expansion.

\subsection{Component and design-space ablations}
\label{sec:supp-components}

\begin{table}[t]
\centering
\small
\setlength{\tabcolsep}{0pt}
\renewcommand{\arraystretch}{1.12}
\caption{\textbf{Stage-2 component and anchoring ablations} (success, \%) ($10$ old and $7$ new tasks on RoboTwin; $10$ old and $20$ new tasks on RoboCasa; $20$ scenes per task).}
\label{tab:components}
\begin{tabular*}{\columnwidth}{@{\extracolsep{\fill}}lccc@{}}
\toprule
Setting & Old & New & All \\
\midrule
\multicolumn{4}{@{}l}{\emph{RoboTwin: expanded-expert velocity anchoring (frozen $f_\phi$)}} \\
With velocity anchoring & \textbf{60.0} & \textbf{47.9} & \textbf{55.0} \\
w/o velocity anchoring & 49.0 & 45.7 & 47.6 \\
\midrule
\multicolumn{4}{@{}l}{\emph{RoboCasa: output anchoring strength}} \\
$\lambda_{\mathrm{keep}}=1.0$ & \textbf{58.5} & 33.0 & 41.5 \\
$\lambda_{\mathrm{keep}}=0.125$ & 57.0 & \textbf{37.8} & \textbf{44.2} \\
\bottomrule
\end{tabular*}
\end{table}

\paragraph{Replay velocity anchoring on expanded experts.}
Under the ``New experts only'' configuration in \cref{tab:plasticity}, where $f_\phi$, $g_\theta$, and inherited S1 experts remain frozen on RoboTwin S2 (\cref{tab:components}, block~1), supervising newly added experts to match the parent velocity on replay data improves old-task retention by $11.0\%$ ($49.0\%\to 60.0\%$) and overall success by $7.4\%$ ($47.6\%\to 55.0\%$). This fallback regularization prevents severe degradation when replay states are routed to an expanded branch.

\paragraph{Output anchoring strength during composite absorption.}
During pure absorption on RoboCasa S2 (\cref{tab:components}, block~2), moderating output velocity anchoring from $\lambda_{\mathrm{keep}}{=}1.0$ to $0.125$ alongside $\mathcal{L}_{\mathrm{feature}}$ improves new-task acquisition by $4.8\%$ ($33.0\%\to 37.8\%$) and raises overall success from $41.5\%$ to $44.2\%$ while retaining $57.0\%$ on prior tasks, giving absorbed specialists sufficient flexibility to smooth multi-primitive transitions while mitigating forgetting.

\section{Row groups of the routing analysis}
\label{sec:supp-families}

The left panel of \Cref{fig:primitive-reuse} groups atomic demonstrations by primitive skill family ($F_1$--$F_{10}$); \cref{tab:primitive-families} lists each family name and window count alongside the composite task groups in the right panel. Each window corresponds to one $50$-frame action chunk governed by a single chunk-level routing decision. Atomic windows are sampled across $3$ episodes per S1 task (stride $20$) and $5$ episodes per S2 task (stride $10$), totaling $406$ and $474$ windows, respectively. Composite windows are sampled across $5$ episodes for each of the $16$ S3 composite tasks (stride $10$) and assigned to the majority primitive label within the chunk. Unseen pouring and hanging motions ($F_{11}$, $F_{12}$) appear exclusively inside the ``new'' composite row because no atomic training task exercises them.

\begin{table}[t]
\centering
\small
\setlength{\tabcolsep}{0pt}
\renewcommand{\arraystretch}{1.12}
\caption{\textbf{Row groups of \cref{fig:primitive-reuse}.} Atomic primitive families $F_k$ of the left panel with their window counts, and the composite task families of the right panel. Each window is a $50$-frame action chunk assigned to its majority primitive label.}
\label{tab:primitive-families}
\begin{tabular*}{\columnwidth}{@{\extracolsep{\fill}}llr@{}}
\toprule
\multicolumn{3}{@{}l}{\emph{Atomic primitive families (left panel)}} \\
\midrule
$F_1$ & single-arm pick-and-place & 222 \\
$F_2$ & reorientation & 34 \\
$F_3$ & pressing & 36 \\
$F_4$ & articulation & 93 \\
$F_5$ & shaking & 223 \\
$F_6$ & tool contact & 86 \\
$F_7$ & dragging & 35 \\
$F_8$ & bimanual lifting & 21 \\
$F_9$ & bimanual parallel grasping & 41 \\
$F_{10}$ & handover & 89 \\
\midrule
\multicolumn{3}{@{}l}{\emph{Composite task families (right panel)}} \\
\midrule
 & pick place & 1{,}216 \\
 & articulated & 92 \\
 & tool & 17 \\
 & bimanual & 600 \\
 & new (including $F_{11}$ pouring, $F_{12}$ hanging) & 163 \\
\bottomrule
\end{tabular*}
\end{table}

\section{RoboTwin continual benchmark composition}
\label{sec:supp-tasks}

Our RoboTwin benchmark assigns all $50$ RoboTwin~2.0 tasks to four cohorts by interaction-unit count and primitive family, parsed from the official expert scripts; \cref{tab:robotwin-tasks} lists the complete division. S1 contains $23$ atomic tasks spanning single-arm pick-and-place, reorientation, pressing, articulation, and bimanual lifting. S2 introduces $7$ tasks whose core primitives are absent from S1 (shaking, tool contact, dragging, bimanual parallel grasping, and handover). S3 adds $16$ multi-step compositions of previously learned primitives, two of which (\texttt{dump\_bin\_bigbin} and \texttt{hanging\_mug}) contain unseen pouring and hanging motions. The remaining $4$ tasks are structurally matched siblings of S2--S3 training tasks (identical composition with different objects or counts) and are held out for zero-shot evaluation only. For RoboCasa, we follow the official lifelong-learning protocol: S1 comprises $65$ atomic tasks, and S2--S4 introduce $20$ composite tasks each ($125$ total).

\begin{table}[t]
\centering
\small
\setlength{\tabcolsep}{4pt}
\renewcommand{\arraystretch}{1.12}
\caption{\textbf{Complete $50$-task division of the RoboTwin continual benchmark.}}
\label{tab:robotwin-tasks}
\begin{tabular}{@{}l p{0.74\columnwidth}@{}}
\toprule
Cohort & Tasks \\
\midrule
S1 (23) & move\_can\_pot, move\_pillbottle\_pad, move\_stapler\_pad, place\_a2b\_left, place\_a2b\_right, place\_container\_plate, place\_empty\_cup, place\_fan, place\_mouse\_pad, place\_object\_scale, place\_object\_stand, place\_phone\_stand, place\_shoe, click\_bell, click\_alarmclock, press\_stapler, turn\_switch, open\_laptop, open\_microwave, rotate\_qrcode, adjust\_bottle, grab\_roller, lift\_pot \\
S2 (7) & shake\_bottle, shake\_bottle\_horizontally, beat\_block\_hammer, stamp\_seal, move\_playingcard\_away, pick\_dual\_bottles, handover\_mic \\
S3 (16) & stack\_blocks\_two, stack\_bowls\_two, stack\_blocks\_three, blocks\_ranking\_rgb, place\_burger\_fries, place\_cans\_plasticbox, place\_dual\_shoes, place\_bread\_basket, place\_bread\_skillet, place\_can\_basket, put\_object\_cabinet, handover\_block, put\_bottles\_dustbin, scan\_object, dump\_bin\_bigbin, hanging\_mug \\
Held-out (4) & place\_object\_basket, stack\_bowls\_three, blocks\_ranking\_size, pick\_diverse\_bottles \\
\bottomrule
\end{tabular}
\end{table}

\section{Baseline implementations and evaluation protocols}
\label{sec:supp-baselines}

\paragraph{Continual-learning baselines.}
Seq.\ FT updates all parameters solely on incoming demonstrations at each stage. RETAIN~\cite{yadav2026retain} linearly interpolates between the previous-stage policy and a newly fine-tuned checkpoint; we sweep its interpolation coefficient over $\{0.25,0.5,0.75\}$, selecting $0.25$ for RoboTwin and $0.5$ for RoboCasa. CLARE~\cite{romer2026clare} freezes the initial $\pi_{0.5}$ policy, attaches stage-specific bottleneck adapters, and selects adapters at inference via autoencoder reconstruction error without storing replay trajectories.

\paragraph{Matched-replay baselines.}
ER+KD updates the single $\pi_{0.5}$ backbone and action branch while distilling the frozen parent policy's velocity predictions on replay batches under the exact per-stage demonstrations, replay ratio $\rho$, and optimization steps used by \method. Dense scales the feedforward width of the single action branch, widening it by $7{,}808$ hidden units in each subsequent stage: on RoboTwin this matches the total parameter count of our expert library at every stage, and on RoboCasa, where \method adds no experts, Dense carries up to $1.3$B more parameters (\cref{tab:params}). FFN-MoE replaces action-transformer feedforward blocks with one shared expert and eight routed experts under top-$2$ gating, appending experts across stages. Both Dense and FFN-MoE train from S1 and apply replay velocity distillation in subsequent stages. Despite their equal or larger capacity, Dense reaches $30.8\%$ and $1.7\%$ on the RoboTwin and RoboCasa S3 cohorts versus $50.9\%$ and $18.3\%$ for \method (\cref{tab:acquisition}).

\begin{table}[t]
\centering
\small
\setlength{\tabcolsep}{3pt}
\renewcommand{\arraystretch}{1.12}
\caption{\textbf{Total parameters} (billions) at each training stage. \method uses $K{=}4$, $6$, $7$ experts in RoboTwin S1--S3 and $K{=}4$ throughout RoboCasa.}
\label{tab:params}
\begin{tabular}{@{}lccc@{\hspace{8pt}}cccc@{}}
\toprule
& \multicolumn{3}{c}{RoboTwin} & \multicolumn{4}{c}{RoboCasa} \\
\cmidrule(lr){2-4}\cmidrule(l){5-8}
Method & S1 & S2 & S3 & S1 & S2 & S3 & S4 \\
\midrule
ER+KD & 3.35 & 3.35 & 3.35 & 3.35 & 3.35 & 3.35 & 3.35 \\
Dense & 5.07 & 5.94 & 6.37 & 5.07 & 5.51 & 5.94 & 6.37 \\
FFN-MoE & 5.17 & 5.62 & 6.07 & 5.17 & 5.62 & 6.07 & 6.52 \\
\method & 5.07 & 5.94 & 6.37 & 5.07 & 5.07 & 5.07 & 5.07 \\
\bottomrule
\end{tabular}
\end{table}

\paragraph{Evaluation protocols.}
All simulated benchmarks report mean success rates across three evaluation seeds under the full protocol: $50$ fixed scenes per task on RoboTwin and $30$ scenes per task on RoboCasa, executing $5$ actions per $50$-step predicted chunk on RoboCasa. The ablation tables in \cref{tab:plasticity,tab:components} evaluate $10$ S1 and $7$ S2 RoboTwin tasks over $20$ scenes each at a $65\%$ episode step horizon, and $10$ old and $20$ new RoboCasa tasks over $20$ scenes each, under the same three evaluation seeds.

\section{Additional implementation details}
\label{sec:supp-implementation}

Across all stages, \method uses batch size $128$, chunk length $H{=}50$, $500$ linear warm-up steps, global gradient norm clipping at $1.0$, exponential moving average (EMA) decay $0.99$, and learning rates of $10^{-4}$ for newly added experts, $10^{-3}$ for the router, and $2.5\times10^{-5}$ for the shared backbone, generalist branch, and unfrozen matched experts. Kinematic motion clustering extracts standardized end-effector translation, cumulative rotation, active-arm masks, and gripper open-close profiles over trajectory segments to automatically determine primitive cluster coverage and expert growth ($M_s$) across stages.

On RoboTwin ($\rho{\approx}4.7\%$ replay; $40$k, $20$k, and $30$k steps across S1--S3), demonstrations are segmented at bimanual gripper open-close transitions. Continual stages anchor both velocity outputs and layer-wise expert-facing features ($\lambda_{\mathrm{keep}}{=}\lambda_{\mathrm{feature}}{=}\lambda_{\mathrm{route}}{=}\lambda_{\mathrm{ret}}{=}1$) while freezing prior router columns. Specifically, S2 freezes all S1 experts ($\mathcal{I}_{\mathrm{abs}}{=}\emptyset$), whereas S3 jointly refines matched specialists ($\mathcal{I}_{\mathrm{abs}}$) and $g_\theta$ alongside the newly added expert while keeping unmatched prior experts frozen.

On RoboCasa ($\rho{=}12.5\%$ replay; $40$k steps per stage across $16$ H20 GPUs), automatic kinematic novelty detection classifies S2--S4 sliding-window chunks within the four S1 clusters ($M_s{=}0$). Chunk targets are initialized from matched specialists or the frozen prior-stage router and refined via $10\%$-margin error competition across the $K{=}4$ experts, while replay chunks are supervised toward the frozen parent router's top-$1$ expert. Training jointly updates matched specialists and $g_\theta$ under $\mathcal{L}_{\mathrm{act}}+\mathcal{L}_{\mathrm{route}}+\mathcal{L}_{\mathrm{feature}}+0.125\,\mathcal{L}_{\mathrm{keep}}$. At deployment across all benchmarks, the router dispatches a single expert $k^{\star}$ per predicted chunk directly from $(o,l)$ prior to fixed-step Euler integration.

On the G2A real robot ($\rho{=}12.5\%$ replay; $20$k optimization steps per stage with batch size $128$ and seed $42$, with training on $16$ NVIDIA A100 GPUs and inference on a single NVIDIA RTX~4090), training uses the $450$ whole-episode demonstrations behind \cref{fig:realrobot-tasks} at $30$\,Hz with $24$-DoF delta actions, segmented at bimanual gripper open--close transitions. S2 freezes all S1 experts and $g_\theta$ while appending the bimanual expert ($K{=}4\to5$); S3 jointly refines the matched placement specialist and $g_\theta$ alongside the newly added left-arm transport expert ($K{=}5\to6$), with unmatched prior experts and prior router columns frozen. Anchoring weights follow the RoboTwin configuration ($\lambda_{\mathrm{keep}}{=}\lambda_{\mathrm{feature}}{=}\lambda_{\mathrm{route}}{=}\lambda_{\mathrm{ret}}{=}1$).